\documentclass[10pt,twocolumn,letterpaper]{article}

\usepackage[pagenumbers]{cvpr} 

\definecolor{cvprblue}{rgb}{0.21,0.49,0.74}
\usepackage[pagebackref,breaklinks,colorlinks,allcolors=cvprblue]{hyperref}
\usepackage{multirow}
\usepackage{booktabs}
\usepackage{graphicx}
\usepackage{algorithm}
\usepackage{algpseudocode}
\def\confName{CVPR}
\def\confYear{2027}

\title{DiT2MoE: Learnable Non-Exclusive MoEfication for Diffusion Transformers}

\author{
Weiyu Huang \qquad
Ziteng Wang \qquad
Jun Zhu \qquad
Jianfei Chen\thanks{
Correspondence to: Jianfei Chen.
}\\[0.5em]
Department of Computer Science and Technology\\
Tsinghua University\\
{\tt\small
\{hwy23,wangzite23\}@mails.tsinghua.edu.cn;
\{dcszj,jianfeic\}@tsinghua.edu.cn
}
}
\begin{document}
\maketitle

\begin{abstract}
Diffusion transformers achieve impressive generation quality, but their high inference costs limit efficient deployment.
MoEfication offers a promising approach to reducing per-step computation by converting pretrained dense FFNs into sparse mixture-of-experts (MoE) layers.
However, existing MoEfication methods often impose restrictive expert assignments and rely on reconstruction-oriented distillation, limiting architectural flexibility and slowing generation-quality recovery.
To address these limitations, we propose DiT2MoE, a two-stage prompt-only distillation framework for MoEfication of diffusion transformers.
Stage 1 jointly learns non-exclusive expert assignments and content-based routing, while introducing lightweight timestep modulation to adjust the contributions of selected experts.
This process uses velocity reconstruction with frozen FFN weights to provide a faithful sparse initialization.
Building on this initialization, Stage 2 accelerates generation-quality recovery through distribution matching and perceptual supervision on student-generated trajectories.
On Qwen-Image-2512 with a 4-step Lightning sampler, DiT2MoE retains 94.9\%--99.8\% of the dense model's scores across GenEval, DPG-Bench, CLIP, and ImageReward, while activating only 37.5\% of the converted FFN weights per token with approximately 400 A800 GPU-hours of training.
We further demonstrate end-to-end inference speedups in realistic deployment settings with an optimized expert kernel.
\end{abstract}

\section{Introduction}
\label{sec:intro}

Diffusion transformers (DiTs) have demonstrated remarkable capabilities in high-quality image synthesis, with performance improving as model capacity and computational budgets increase~\cite{peebles2023dit,chen2024pixart,esser2024scaling}.
Despite these advances, deploying large-scale DiTs remains computationally expensive, as image synthesis requires repeated evaluations of a large transformer.
While recent few-step distillation methods significantly shorten the denoising process~\cite{yin2024dmd,yin2024dmd2}, each denoising step still incurs the full computational cost of the dense model.
This motivates exploring more efficient architectures that reduce per-step computation without sacrificing the generation capabilities of pretrained models.

MoEfication offers a promising way to reduce
inference costs by converting pretrained dense
FFNs into sparse experts, which involves assigning FFN neurons
to experts and initializing the router weight.
Early LLM MoEfication methods construct fixed
expert partitions through activation-based grouping
~\cite{zhang2022moefication,zhu2024llamamoe,
lee2024breaking,pei2026analytical}
or exploit activation sparsity
~\cite{szatkowski2024exploiting}.
Dense2MoE~\cite{zheng2025dense2moe} extends
this paradigm to diffusion transformers using
fixed Taylor-based expert partitions followed by
image-text distillation.
However, fixed partitions offer limited
structural adaptability, motivating approaches ~\cite{feng2025dive,gao2026tomoe,bamba2026dotmoe}
that learn neuron-expert assignments through pruning
and joint optimization.

Despite progress in learnable expert construction,
existing methods largely retain disjoint assignments.
This limitation is particularly relevant to DiTs,
where neuron contributions are highly nonuniform,
with some neurons consistently contributing
much more than others.
Disjoint assignments restrict each neuron to a
single expert, forcing less useful neurons to occupy
expert capacity while preventing important neurons
from serving multiple routing contexts,
leading to suboptimal sparse initialization
and costly quality recovery.
Furthermore, neuron contributions vary across
denoising steps, complicating sparse conversion.
Existing diffusion MoEfication methods also
often require real-image supervision, while
reconstruction-based objectives do not directly
optimize generation quality.

To address these challenges, we propose DiT2MoE, a two-stage framework for prompt-only dense-to-MoE conversion.
In Stage 1, we construct a flexible sparse structure through non-exclusive expert assignments and content-based routing, complemented by lightweight timestep modulation of the selected expert outputs. We reserve high-importance neurons for shared experts and jointly optimize the remaining assignments with a scaled Top-K router. Using teacher-generated trajectories, Stage 1 learns sparse assignments and routing through velocity reconstruction with frozen FFN weights, reducing the initial prediction gap. Stage 2 then fixes the learned structure and fine-tunes the FFN weights and router with on-policy distillation, combining distribution matching with perceptual supervision to recover generation quality beyond pointwise velocity reconstruction.
Experiments demonstrate that DiT2MoE substantially outperforms existing MoEfication methods on various diffusion models in generation quality under matched FFN activation budgets, while requiring no real training images and incurring only a modest training cost. DiT2MoE largely preserves dense-model quality with substantially reduced computation, and achieves up to 1.25x end-to-end speedup using an optimized expert kernel.

\section{Preliminaries}
\label{sec:preliminaries}

We provide a mathematical formulation for few-step diffusion sampling and FFN MoEfication,
then present empirical observations motivating our sparse expert design.

\subsection{Diffusion Formulation and Sampling}
\label{sec:preliminaries_dit}

In flow-matching diffusion transformers (DiTs), the model
predicts a velocity $v_{\theta}(\mathbf{z}_{\sigma},\sigma,c)$,
where $\mathbf{z}_{\sigma}$ is the noisy latent,
$\sigma\in[0,1]$ is the noise level, and $c$ is the
text condition.
Starting from $\mathbf{z}_{1}\sim\mathcal{N}(\mathbf{0},\mathbf{I})$,
sampling follows a decreasing noise schedule
$1=\sigma_0>\cdots>\sigma_N=0$ with Euler updates:
\begin{equation}
    \mathbf{z}_{\sigma_{i+1}}
    =
    \mathbf{z}_{\sigma_i}
    +(\sigma_{i+1}-\sigma_i)
    v_{\theta}(\mathbf{z}_{\sigma_i},\sigma_i,c).
    \label{eq:flow_sampling}
\end{equation}
Under the linear flow formulation
$\mathbf{z}_{\sigma}=(1-\sigma)\mathbf{z}_0
+\sigma\boldsymbol{\epsilon}$, the predicted clean latent is
\begin{equation}
    \widehat{\mathbf{z}}_0
    =
    \mathbf{z}_{\sigma}
    -\sigma v_{\theta}(\mathbf{z}_{\sigma},\sigma,c).
    \label{eq:clean_prediction}
\end{equation}
We denote the sampling process as
$\mathbf{z}_0^{\theta}=G_{\theta}(\boldsymbol{\epsilon},c)$
and its intermediate states as a denoising rollout.
We focus on pretrained few-step diffusion models and
preserve their original sampling schedules.

\subsection{Dense FFNs and MoEfication}
\label{sec:preliminaries_moe}

For an input token $\mathbf{x}\in\mathbb{R}^{d}$,
a dense FFN computes
\begin{equation}
    F(\mathbf{x})
    =
    \mathbf{W}_{\mathrm{down}}\mathbf{h}(\mathbf{x})
    +\mathbf{b}_{\mathrm{down}},
    \label{eq:dense_ffn}
\end{equation}
where $\mathbf{h}(\mathbf{x})\in\mathbb{R}^{D}$
is the intermediate activation.
For standard FFNs,
$\mathbf{h}(\mathbf{x})
=\phi(\mathbf{W}_{\mathrm{up}}\mathbf{x}
+\mathbf{b}_{\mathrm{up}})$,
while gated FFNs additionally apply element-wise gating.
We omit layer indices for clarity.

Conventional partition-based MoEfication divides the
$D$ neurons into $E$ equal-sized experts, each containing
$m=D/E$ neurons.
With $S$ always-active shared experts and $E-S$ routed
experts, the router selects $K$ experts per token:
\begin{equation}
\begin{aligned}
    \mathbf{g}(\mathbf{x})
    &=
    \operatorname{TopKSoftmax}
    (\mathbf{W}_r\mathbf{x}+\mathbf{b}_r,K),\\
    \widetilde F(\mathbf{x})
    &=
    \mathbf{b}_{\mathrm{down}}
    +f_{\mathrm{sh}}(\mathbf{x})
    +\sum_{e=1}^{E-S}g_e(\mathbf{x})f_e(\mathbf{x}),
\end{aligned}
\label{eq:moefication}
\end{equation}
where $f_{\mathrm{sh}}(\mathbf{x})
=\sum_{s=1}^{S}f_s^{\mathrm{sh}}(\mathbf{x})$
aggregates shared expert outputs, and $f_e(\mathbf{x})$
denotes the output of routed expert $e$.
$\operatorname{TopKSoftmax}$ normalizes the selected
$K$ routing weights and sets the others to zero.
Each expert reuses the pretrained FFN weights of its
assigned neurons.

\begin{figure}[t]
    \centering
    \includegraphics[width=1.0\columnwidth]{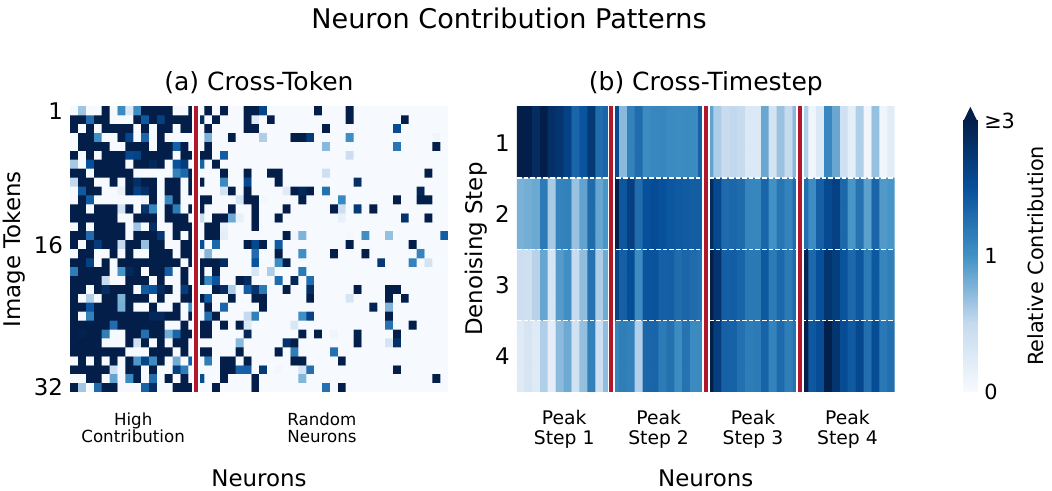}
    \caption{
    Neuron contribution patterns in the image FFN of
    Qwen-Image-2512 (block 29).
    \textbf{Left (cross-token):}
    Contributions of 16 neurons sampled from the top quartile by average contribution and
    32 randomly sampled neurons across eight calibration prompts at denoising step 2.
    \textbf{Right (cross-timestep):}
Contributions of top-quartile candidate neurons at each timestep exhibiting pronounced variations across denoising steps.
    Colors are relative to the all-neuron mean per token
    (left) or timestep (right).
    }
    \label{fig:neuron_contribution}
\end{figure}

\subsection{Neuron Contribution Patterns in Diffusion Transformers}
\label{sec:empirical}

We analyze neuron contributions in Qwen-Image-2512 using
sampled text prompts and a 4-step Lightning sampler.
The contribution of neuron $j$ at denoising step $b$
is measured as
\begin{equation}
    q_j^{(b)}
    =
    \frac{1}{|\mathcal{X}_b|}
    \sum_{\mathbf{x}\in\mathcal{X}_b}
    |h_j(\mathbf{x})|
    \|\mathbf{W}_{\mathrm{down}}[:,j]\|_2,
    \label{eq:neuron_contribution}
\end{equation}
where $\mathcal{X}_b$ denotes the collected image tokens
at step $b$, and $h_j$ is the activation of neuron $j$.

As shown in Fig.~\ref{fig:neuron_contribution},
we investigate neuron contributions across image
tokens and denoising steps.
For the cross-token analysis, we sample neurons
from the top quartile by average contribution
and compare them with randomly sampled neurons.
The former consistently exhibit much higher
contributions across tokens, revealing a pronounced
imbalance in neuron importance.
Such uneven contributions suggest that important
neurons may be useful across multiple routing
contexts, while less useful neurons contribute
relatively little.
Disjoint expert partitions may therefore
underutilize important neurons while allocating
capacity to less useful ones.
For the cross-timestep analysis, we visualize
neurons with pronounced contribution variations
across denoising steps.
Their contribution magnitudes change substantially
with noise levels, indicating that neuron importance
is not constant throughout sampling.
These observations highlight the
uneven distribution of neuron importance and
its variation across denoising stages.


\begin{figure*}[t]
    \centering
    \includegraphics[width=\textwidth]{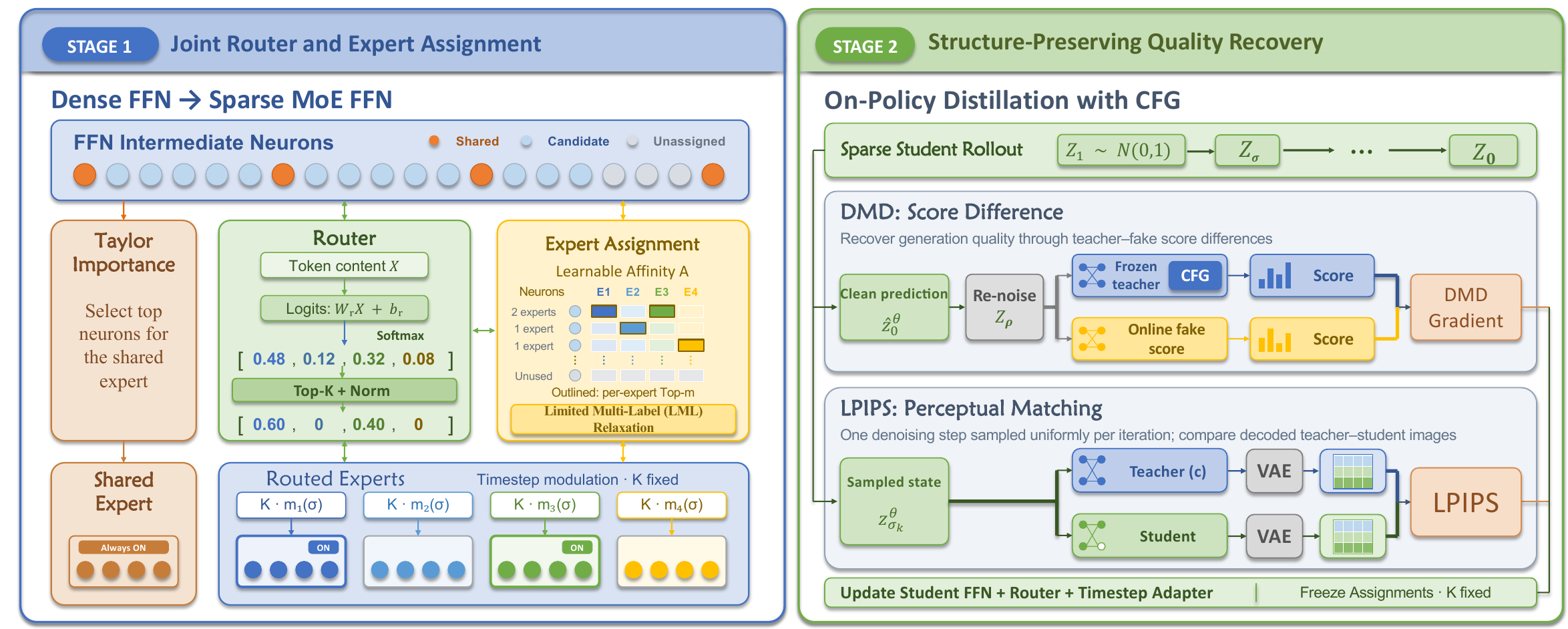}
    \caption{
        Overview of DiT2MoE.
        Stage 1 efficiently initializes the sparse model through joint expert assignment and routing optimization with frozen FFN weights. Stage 2 fixes the learned sparse structure and recovers generation quality through prompt-only distribution matching and perceptual supervision.
    }
    \label{fig:overview}
\end{figure*}

\section{Method}
\label{sec:method}

In this section, we present DiT2MoE, a two-stage framework for
converting pretrained diffusion transformers into
sparse MoE models using only text prompts.
Section~\ref{sec:stage1} establishes the sparse
architecture through teacher-velocity reconstruction,
while Section~\ref{sec:stage2} fixes its structure
and recovers generation quality through
student-trajectory distillation. The complete two-stage training procedure is summarized in Appendix A.

\subsection{Stage 1: Joint Routing and Expert Assignment Optimization}
\label{sec:stage1}

Direct sparse conversion can introduce substantial
prediction errors, making subsequent generative
distillation challenging.
To obtain a faithful initialization, we jointly optimize expert assignments, routing, and timestep modulation through lightweight teacher-velocity reconstruction, while keeping pretrained FFN weights frozen.

\paragraph{Learnable Expert Construction}
\label{sec:stage1_partition}
There are two main challenges in expert construction: unequal neuron
importance and complex neuron grouping.
As observed in our analysis, neuron contributions vary substantially across tokens, with some neurons consistently exhibiting high contributions whereas others contribute less on average.
We therefore select consistently important neurons using Taylor importance to construct $S$ always-active shared experts. Section B of the supplementary material details the Taylor scoring procedure.
Moreover, determining how to group the remaining neurons
is nontrivial, as their usefulness depends on different
routing contexts, requiring expert assignments to
adapt jointly with the router during training.
Disjoint partitions force less useful neurons to occupy expert capacity while preventing useful neurons from serving different routing contexts. We therefore learn independent, capacity-constrained assignments jointly with routing, allowing selective neuron reuse without requiring the corresponding experts to be activated together.

Specifically, after fixing the shared neurons, we consider
the remaining $N=D-Sm$ neurons and $R=E-S$ routed experts,
each with capacity $m=D/E$.
Our goal is to learn a binary assignment matrix
$\mathbf{M}\in\{0,1\}^{N\times R}$,
where $M_{je}$ indicates whether neuron $j$ belongs
to expert $e$.
We require $\sum_j M_{je}=m$ for each expert,
without constraining the row sums.
This allows neurons to appear in multiple experts
or remain unassigned while preserving expert capacity.

Since the discrete matrix $\mathbf{M}$ cannot be
directly optimized through gradient descent,
we introduce a learnable affinity matrix
$\mathbf{A}\in\mathbb{R}^{N\times R}$,
where $A_{je}$ represents the assignment score
of neuron $j$ to expert $e$.
We use a differentiable projection to approximate the desired $m$-hot assignments. Specifically,
we construct a soft assignment
$\mathbf{P}\in[0,1]^{N\times R}$
using the Limited Multi-Label (LML)
relaxation~\cite{amos2019limited}:

\begin{equation}
    P_{je}
    =
    \operatorname{sigmoid}
    \left(\frac{A_{je}-\lambda_e}{\tau}\right),
    \qquad
    \sum_j P_{je}=m,
    \label{eq:lml_assignment}
\end{equation}
where $P_{je}$ represents the soft membership
score of neuron $j$ in expert $e$,
$\tau>0$ controls assignment softness, and
$\lambda_e$ is an affinity-dependent threshold
enforcing the capacity constraint.
This relaxation preserves the per-expert capacity
and, unlike an $m$-scaled softmax, bounds individual
membership scores within $[0,1]$.
While scaled softmax concentrates its mass on a single
neuron as $\tau\to0$, LML approaches an $m$-hot
assignment, assuming distinct scores at the
selection boundary.
During the forward pass, each expert independently
selects its top-$m$ neurons:
\begin{equation}
    M_{je}
    =
    \mathbb{I}\left[
    j\in\operatorname{Top}_m(\mathbf{A}_{:,e})
    \right]
    =
    \mathbb{I}\left[
    j\in\operatorname{Top}_m(\mathbf{P}_{:,e})
    \right].
    \label{eq:hard_assignment}
\end{equation}
The two selections are equivalent because LML
preserves the ranking of affinity scores within
each expert.

To propagate gradients through the discrete
assignments, we employ a straight-through estimator:
\begin{equation}
    \widetilde{\mathbf{M}}
    =
    \mathbf{P}
    +\operatorname{sg}(\mathbf{M}-\mathbf{P}),
    \label{eq:partition_ste}
\end{equation}
where $\operatorname{sg}(\cdot)$ denotes stop-gradient.
The forward pass uses hard assignments,
while the backward pass propagates gradients
through $\mathbf{P}$, enabling optimization
of expert assignments. Further details are provided in Appendix B.

\paragraph{Scaled Top-$K$ Routing and Timestep Modulation}
\label{sec:stage1_router}

Unlike conventional MoE layers, pretrained dense FFNs
are not trained with normalized routing weights.
Applying Top-$K$ softmax thus attenuates their original
contributions. We compensate with a fixed scaling factor
$\alpha=K$, since selected routing weights average $1/K$.
Although pretrained DiTs already employ timestep
modulation on attention and FFN outputs, these
mechanisms operate at the branch level rather than
on individual routed experts.
We therefore preserve the original modulation
and introduce an additional expert-wise factor
$m_e(\sigma)$ to adapt expert contributions
across denoising steps, as motivated by
Sec.~\ref{sec:empirical}.

\begin{equation}
    m_e(\sigma)
    =
    1+\mathbf{v}_e^\top
    \operatorname{ReLU}
    \left(
    \mathbf{U}\operatorname{Emb}(\sigma)
    \right),
    \label{eq:timestep_adapter}
\end{equation}
where $\operatorname{Emb}(\sigma)\in\mathbb{R}^d$
is a sinusoidal noise-level embedding,
$\mathbf{U}\in\mathbb{R}^{r\times d}$ is a shared
low-rank projection, and $\mathbf{v}_e\in\mathbb{R}^r$
is an expert-specific head. We initialize
$\mathbf{v}_e=0$ so that $m_e(\sigma)=1$ initially.
Shared experts receive no additional expert-wise modulation.

Let $\boldsymbol{\ell}(\mathbf{x})
=\mathbf{W}_r\mathbf{x}+\mathbf{b}_r$
denote the content-based router logits, and let
$\mathbf{g}(\mathbf{x})
=\operatorname{TopKSoftmax}
(\boldsymbol{\ell}(\mathbf{x}),K)$
denote the normalized routing weights.
We integrate routing, timestep modulation, and
learned assignments into neuron-level gates:
\begin{equation}
    \mathbf{q}(\mathbf{x},\sigma)
    =
    \widetilde{\mathbf{M}}
    \left(
    K\,\mathbf{m}(\sigma)
    \odot\mathbf{g}(\mathbf{x})
    \right).
    \label{eq:neuron_gate}
\end{equation}
The timestep modulation rescales the contributions of already-selected experts, while the Top-K selection remains determined by content-based routing logits. The assignment matrix maps these modulated expert weights to neuron-level gates, aggregating contributions from multiple experts for overlapping neurons.
The resulting sparse FFN is
\begin{equation}
\begin{aligned}
    \widetilde{F}(\mathbf{x},\sigma)
    &=
    \mathbf{b}_{\mathrm{down}}
    +f_{\mathrm{sh}}(\mathbf{x})\\
    &\quad+
    \mathbf{W}_{\mathrm{down,rt}}
    \left(
    \mathbf{h}_{\mathrm{rt}}(\mathbf{x})
    \odot\mathbf{q}(\mathbf{x},\sigma)
    \right),
\end{aligned}
\label{eq:moe_forward}
\end{equation}
where $f_{\mathrm{sh}}$ is the shared expert output,
and $\mathbf{h}_{\mathrm{rt}}$ and
$\mathbf{W}_{\mathrm{down,rt}}$ are the non-shared
activations and down-projection weights.

At inference, we materialize the hard expert
assignments and absorb $K$ into the expert
down-projection weights.

\paragraph{Joint Training with Teacher Rollouts}
\label{sec:stage1_objective}

At the initial stage of sparse conversion,
poorly initialized assignments and routing can
introduce substantial prediction errors.
We therefore use teacher-trajectory velocity MSE
to provide stable and inexpensive supervision
for learning a faithful sparse initialization.
Unlike previous methods relying on real image-text
data, we sample text prompts $c$ and Gaussian noise
$\mathbf{z}_1\sim\mathcal{N}(\mathbf{0},\mathbf{I})$,
then run the pretrained teacher to collect
intermediate denoising states $\mathbf{z}_\sigma$.
Given the resulting rollout samples
$(\mathbf{z}_\sigma,\sigma,c)
\sim\mathcal{D}_{\mathrm{T}}$,
we minimize
\begin{equation}
    \mathcal{L}_{\mathrm{MSE}}
    =
    \mathbb{E}_{\mathcal{D}_{\mathrm{T}}}
    \left[
    \left\|
    v_{\theta}(\mathbf{z}_{\sigma},\sigma,c)
    -
    \operatorname{sg}\left(
    v_{\mathrm{T}}(\mathbf{z}_{\sigma},\sigma,c)
    \right)
    \right\|_2^2
    \right],
    \label{eq:output_loss}
\end{equation}
where $\mathcal{D}_{\mathrm{T}}$ denotes the
teacher rollout distribution,
$\mathbf{z}_{\sigma}$ is the noisy latent at
noise level $\sigma$, and $c$ is the text condition.
Here, $v_{\theta}$ and $v_{\mathrm{T}}$ are the
student and teacher velocity predictions,
respectively.
We additionally employ standard MoE load-balancing
$\mathcal{L}_{\mathrm{bal}}=R\sum_{e=1}^{R}r_ep_e$
and router z-loss
$\mathcal{L}_z=\frac{1}{B}\sum_{i=1}^{B}
(\log\sum_{e=1}^{R}\exp(\ell_{ie}))^2$
regularization, where $R$ is the number of routed
experts, $B$ is the number of tokens, $r_e$ is
the fraction of Top-$K$ selections assigned to
expert $e$, $p_e$ is its mean pre-Top-$K$ softmax
probability, and $\ell_{ie}$ is the router logit
for token $i$ and expert $e$.
The overall Stage~1 objective is
\begin{equation}
    \mathcal{L}_1
    =
    \mathcal{L}_{\mathrm{MSE}}
    +\lambda_{\mathrm{bal}}\mathcal{L}_{\mathrm{bal}}
    +\lambda_z\mathcal{L}_z,
    \label{eq:stage1_loss}
\end{equation}
where $\lambda_{\mathrm{bal}}$ and $\lambda_z$
are the regularization weights.

\subsection{Stage 2: Distribution and Perceptual Quality Recovery}
\label{sec:stage2}

While Stage 1 recovers teacher velocities on
teacher-generated trajectories, local reconstruction
does not directly optimize generation quality.
In our preliminary experiments, velocity-only
recovery tends to produce overly smoothed images
with blurred fine details, potentially due to
the averaging effect of MSE regression.
Moreover, under few-step sampling, matching
instantaneous teacher velocities does not guarantee
matching the final generation distribution due to
large denoising intervals.
We therefore adopt DMD to directly align the
student's generation distribution with the teacher,
improving overall visual realism without requiring
exact velocity matching.
We further introduce LPIPS to preserve perceptual
details through feature-level supervision.
Both objectives operate on student-generated trajectories, while the learned expert assignments remain fixed. Overlapping neurons are initialized from the same pretrained weights but independently fine-tuned. We update the FFN, routing, and timestep-modulation parameters while preserving the sparse structure.

\paragraph{Distribution Matching}
We follow the backward-simulation strategy of DMD~\cite{yin2024dmd,yin2024dmd2}.
Starting from Gaussian noise
$\mathbf{z}_1\sim\mathcal{N}(\mathbf{0},\mathbf{I})$,
we sample a denoising step $k$, execute the preceding student denoising steps without tracking gradients, and differentiate through the step-k clean-latent prediction $\mathbf{z}_0^\theta$.
We re-noise this prediction as
$\mathbf{z}_\rho^\theta
=(1-\rho)\mathbf{z}_0^\theta+\rho\boldsymbol{\epsilon}$,
where $\boldsymbol{\epsilon}\sim
\mathcal{N}(\mathbf{0},\mathbf{I})$.
A CFG-guided dense teacher $v_{\mathrm{T}}$
and an online unguided fake-score model $v_\psi$
produce clean-latent estimates
$\widehat{\mathbf{z}}_{0,\rho}^{i}
=\mathbf{z}_\rho^\theta
-\rho v_i(\mathbf{z}_\rho^\theta,\rho,c)$,
for $i\in\{\mathrm{T},\psi\}$.
The DMD gradient is
\begin{equation}
    \nabla_\theta\mathcal{L}_{\mathrm{DM}}
    =
    \mathbb{E}\left[
    \left(\frac{\partial\mathbf{z}_0^\theta}
    {\partial\theta}\right)^{\!\top}
    \operatorname{sg}\left(
    \frac{
    \widehat{\mathbf{z}}_{0,\rho}^{\psi}
    -\widehat{\mathbf{z}}_{0,\rho}^{\mathrm{T}}
    }{\nu}
    \right)
    \right],
    \label{eq:dm_gradient}
\end{equation}
where $\nu$ is a detached normalization factor.
The fake-score model is initialized from the dense
teacher and updated online using flow-matching on re-noised student samples.

\paragraph{Perceptual Supervision}
We additionally employ LPIPS~\cite{zhang2018unreasonable}
on paired clean-latent predictions from the same
student-visited state $\mathbf{z}_\sigma^\theta$:
$\widehat{\mathbf{z}}_{0,\sigma}^{i}
=\mathbf{z}_\sigma^\theta
-\sigma v_i(\mathbf{z}_\sigma^\theta,\sigma,c)$,
for $i\in\{\theta,\mathrm{T}\}$.
The perceptual loss is
\begin{equation}
\begin{aligned}
    \mathcal{L}_{\mathrm{LPIPS}}
    =\mathbb{E}_{\mathcal{D}_\theta}\Big[
    \operatorname{LPIPS}\big(
    &\mathcal{C}(\widehat{\mathbf{z}}_{0,\sigma}^{\theta}),\\
    &\mathcal{C}(\operatorname{sg}(
    \widehat{\mathbf{z}}_{0,\sigma}^{\mathrm{T}}))
    \big)\Big],
\end{aligned}
\label{eq:lpips_loss}
\end{equation}
where $\mathcal{D}_\theta$ denotes student trajectories
and $\mathcal{C}$ denotes VAE decoding of the
output latent.
This provides perceptual supervision without
real reference images.

\paragraph{Overall Objective}
The Stage 2 objective integrates distribution matching,
perceptual supervision, and routing regularization:
\begin{equation}
\begin{aligned}
    \mathcal{L}_2
    ={}&\lambda_{\mathrm{DM}}\mathcal{L}_{\mathrm{DM}}
    +\lambda_{\mathrm{LPIPS}}\mathcal{L}_{\mathrm{LPIPS}}\\
    &+\lambda_{\mathrm{bal}}\mathcal{L}_{\mathrm{bal}}
    +\lambda_z\mathcal{L}_z.
\end{aligned}
\label{eq:stage2_loss}
\end{equation}
All supervision is derived from text prompts
and model-generated samples without real images.

\section{Related Work}
\label{sec:related_work}

\paragraph{Efficient Diffusion Inference}
Existing approaches accelerate diffusion models through pruning and quantization. Pruning reduces model complexity by removing redundant U-Net blocks in BK-SDM~\cite{kim2024bksdm} or structures in diffusion transformers~\cite{hu2026ditpruning}. Related advances in model sparsification include hardware-efficient 2:4 sparsity~\cite{hu2024accelerating}, adaptive semi-structured pruning~\cite{huang2025pruning}, continuous differentiable sparsity-aware training~\cite{huang2025cast}, and differentiable structured pruning~\cite{huang2026deterministic}. Quantization further reduces computational and memory costs through low-precision operations~\cite{li2023qdiffusion,li2025svdquant}, with sensitivity-aware techniques improving weight preservation~\cite{hu2025identifying}.

\paragraph{MoEfication of Pretrained Transformers}
MoEfication transforms pretrained dense FFNs into
conditionally activated experts to reduce inference
cost while preserving model capacity.
Early MoEfication~\cite{zhang2022moefication}
exploits sparse neuron activations to partition FFNs
and train expert routers, while
MoEBERT~\cite{zuo2022moebert} adopts
importance-guided expert construction and
layer-wise distillation.
DIVE~\cite{feng2025dive} constructs diverse experts
through domain-specific pruning and reconstruction.
More recently, CMoE~\cite{pei2026analytical}
uses activation statistics to construct shared
and routed experts with analytically derived routing.
DOT-MoE~\cite{bamba2026dotmoe} formulates neuron
partitioning as balanced optimal transport,
jointly optimizing mutually exclusive assignments
and routing through differentiable Sinkhorn
relaxation.
For diffusion transformers,
Dense2MoE~\cite{zheng2025dense2moe}
constructs shared experts using Taylor-based importance, followed by
knowledge distillation and selective block
activation.
Our approach instead learns non-exclusive
expert assignments with content-based routing
and timestep-dependent expert output modulation.

\paragraph{Diffusion Distillation}

Progressive distillation~\cite{salimans2022progressive}
iteratively reduces the required denoising steps,
while consistency models~\cite{song2023consistency}
and latent consistency models~\cite{luo2023lcm}
enable fast sampling through consistency-based
training.
Distribution Matching Distillation
(DMD)~\cite{yin2024dmd,yin2024dmd2} directly aligns
student and teacher generation distributions
through score-based supervision.
Unlike methods that target sampling-step reduction, we employ DMD together with LPIPS-based perceptual supervision ~\cite{zhang2018unreasonable} to recover generation
quality after MoEfication.

\begin{table*}[t]
    \centering
    \caption{
    Generation quality comparison across three
    few-step diffusion transformers.
    All sparse methods activate 37.5\% of the
    converted FFN weights.
    DiT2MoE (Init.) denotes the sparse initialization
    obtained with frozen pretrained FFN weights.
    }
    \label{tab:main_results}
    \scriptsize
    \setlength{\tabcolsep}{4pt}
    \renewcommand{\arraystretch}{1.12}
    \resizebox{\textwidth}{!}{%
    \begin{tabular}{@{}l cccc cccc cccc@{}}
        \toprule
        \multirow{2}{*}{Method}
        & \multicolumn{4}{c}{Qwen-Image-2512}
        & \multicolumn{4}{c}{Qwen-Image-2.1-Turbo}
        & \multicolumn{4}{c}{FLUX.1-schnell} \\
        \cmidrule(lr){2-5}
        \cmidrule(lr){6-9}
        \cmidrule(l){10-13}
        & GenEval $\uparrow$
        & DPG $\uparrow$
        & CLIP $\uparrow$
        & IR $\uparrow$
        & GenEval $\uparrow$
        & DPG $\uparrow$
        & CLIP $\uparrow$
        & IR $\uparrow$
        & GenEval $\uparrow$
        & DPG $\uparrow$
        & CLIP $\uparrow$
        & IR $\uparrow$ \\
        \midrule
        Dense Teacher
        & 0.7560 & 86.11 & 0.2929 & 1.1276
        & 0.6859 & 82.83 & 0.2950 & 0.8923
        & 0.6524 & 84.69 & 0.2923 & 0.9187 \\
        \midrule
    CMoE
    & 0.6819 & 83.89 & \textbf{0.2928} & 0.8530
    & 0.5753 & 75.45 & 0.2912 & 0.5971
    & 0.5394 & 77.12 & 0.2852 & 0.6582 \\
        Dense2MoE
        & 0.6537 & 82.81 & 0.2861 & 0.8111
        & 0.5591 & 73.94 & 0.2813 & 0.5866 
        & 0.5198 & 76.56 & 0.2755 & 0.6207 \\
        \midrule
        DiT2MoE (Init.)
        & 0.6453 & 83.89 & 0.2890 & 0.8813 
        & 0.5402 & 74.28 & 0.2826  & 0.6475  
        & 0.5305 &  78.10& 0.2880 &  0.6667\\
        \textbf{DiT2MoE (Full)}
        & \textbf{0.7399} & \textbf{85.52} & 0.2923 & \textbf{1.0701}
        & \textbf{0.6224} & \textbf{78.72} & \textbf{0.2931} & \textbf{0.8123}
        &  \textbf{0.6079}  & \textbf{82.62}  & \textbf{0.2906}  & \textbf{0.8229}\\
        \bottomrule
    \end{tabular}%
    }
\end{table*}

\begin{figure*}[t]
    \centering
    \setlength{\tabcolsep}{2pt}
    \renewcommand{\arraystretch}{1.0}
    \scriptsize
    \begin{tabular}{@{}ccccc@{}}
        Dense Teacher &
        CMoE &
        Dense2MoE &
        DiT2MoE (Init.) &
        \textbf{\shortstack{DiT2MoE (Full)}} \\[2pt]

        \includegraphics[width=0.192\textwidth]{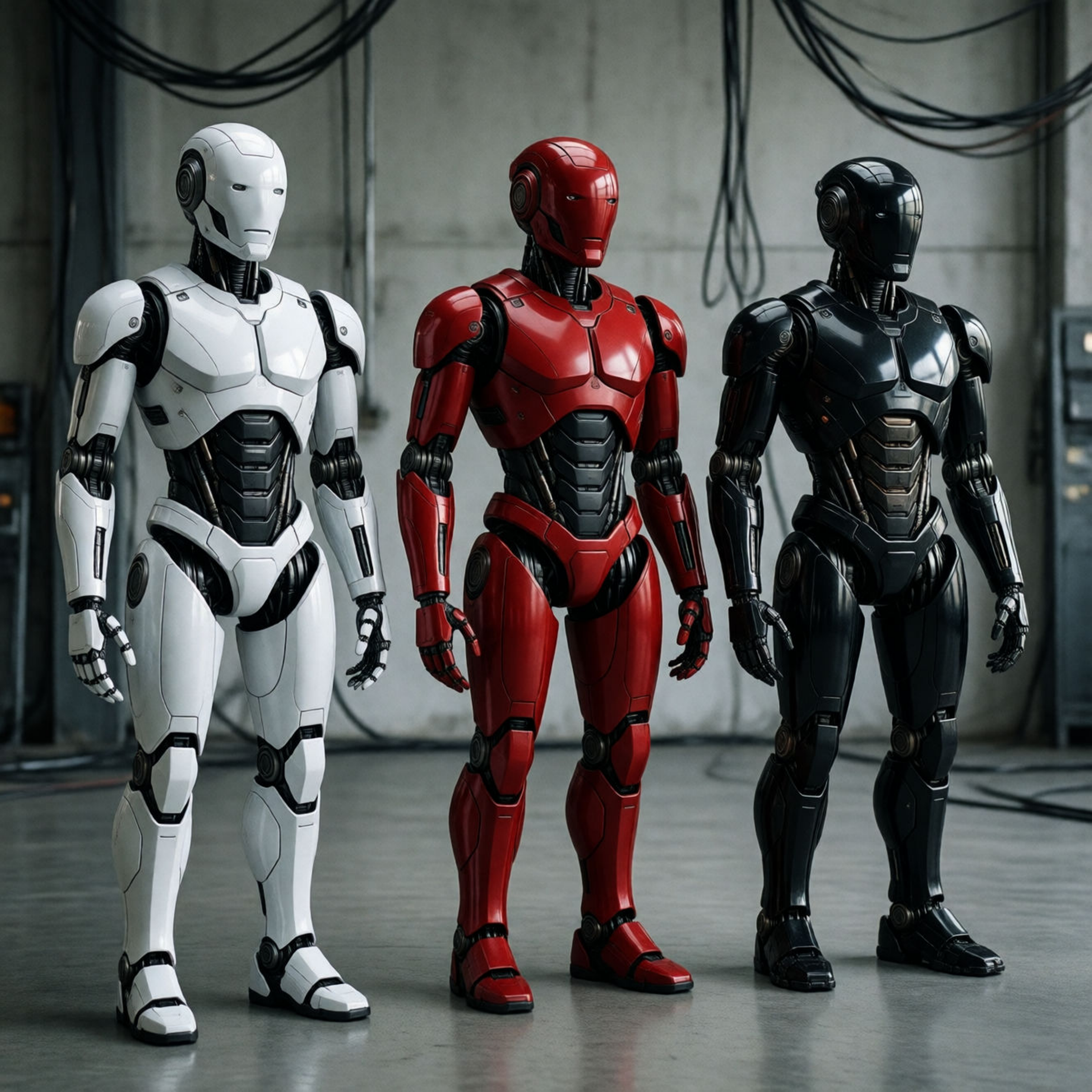} &
        \includegraphics[width=0.192\textwidth]{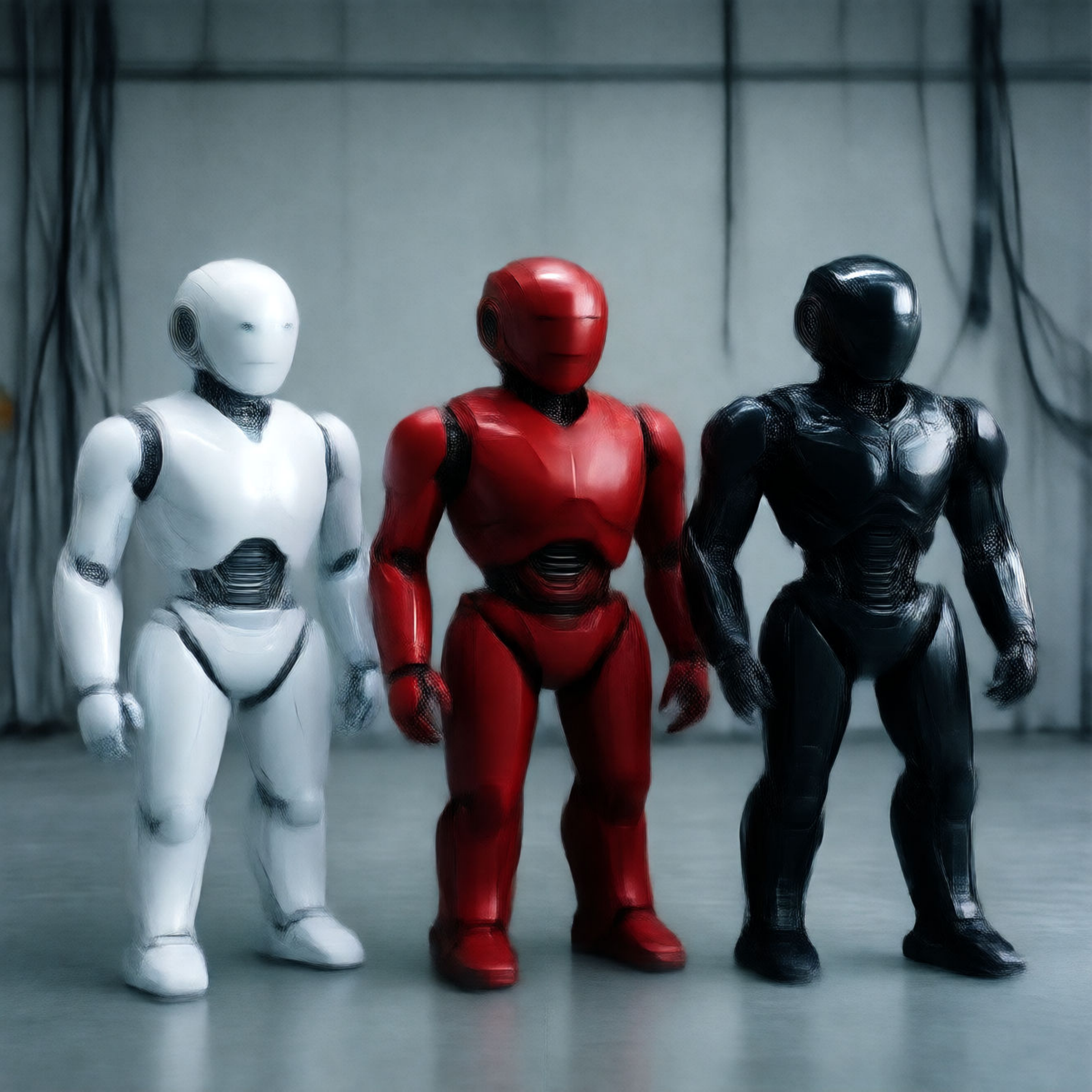} &
        \includegraphics[width=0.192\textwidth]{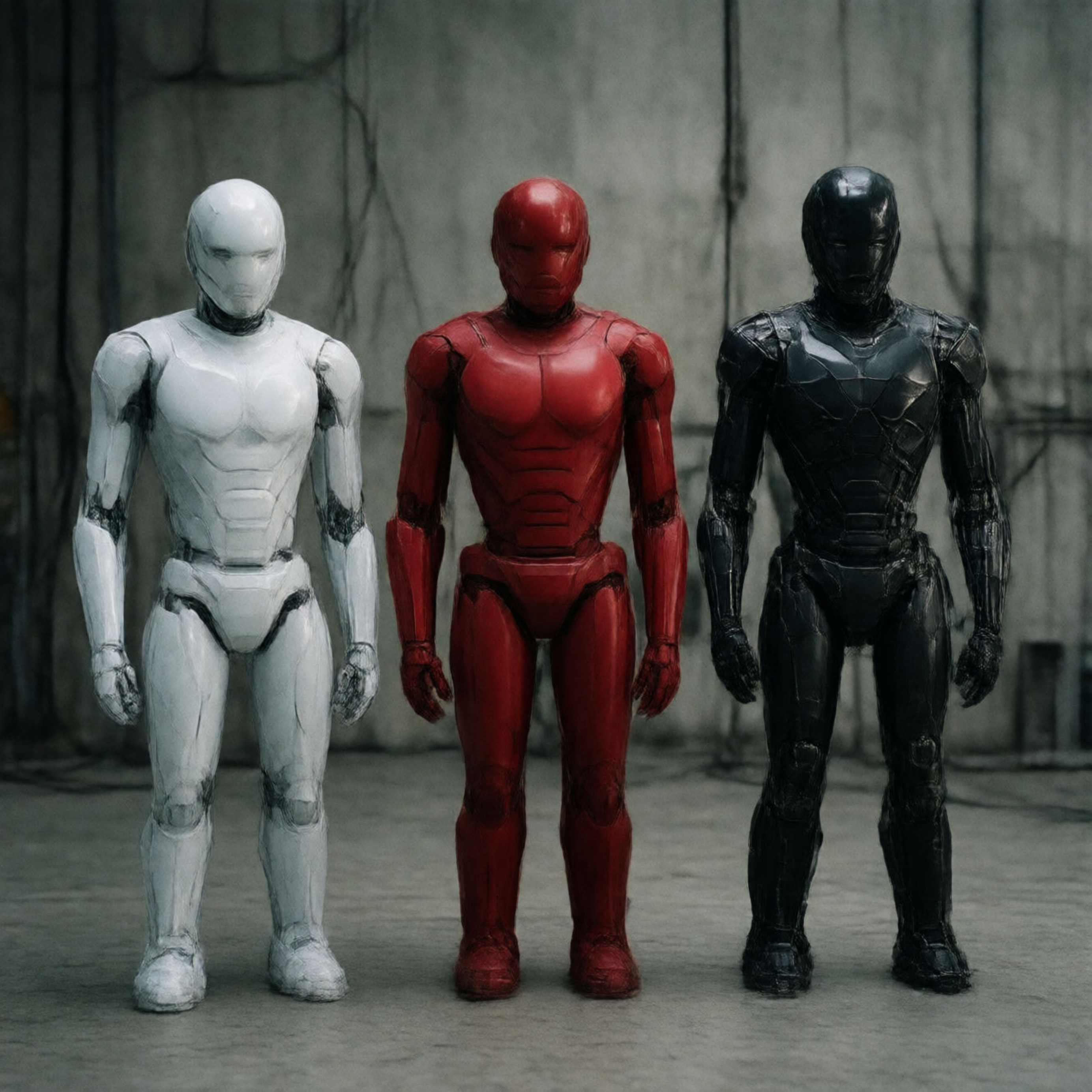} &
        \includegraphics[width=0.192\textwidth]{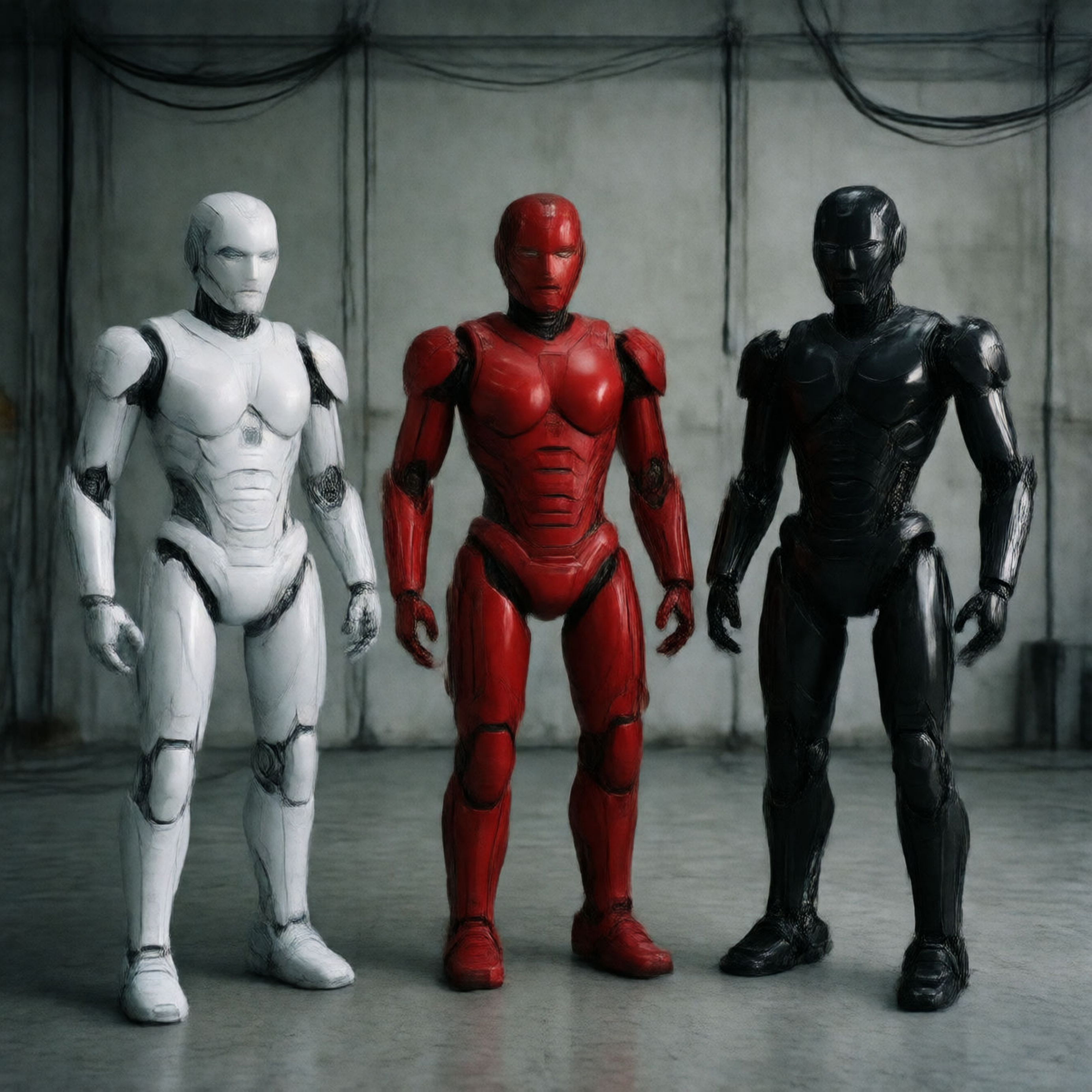} &
        \includegraphics[width=0.192\textwidth]{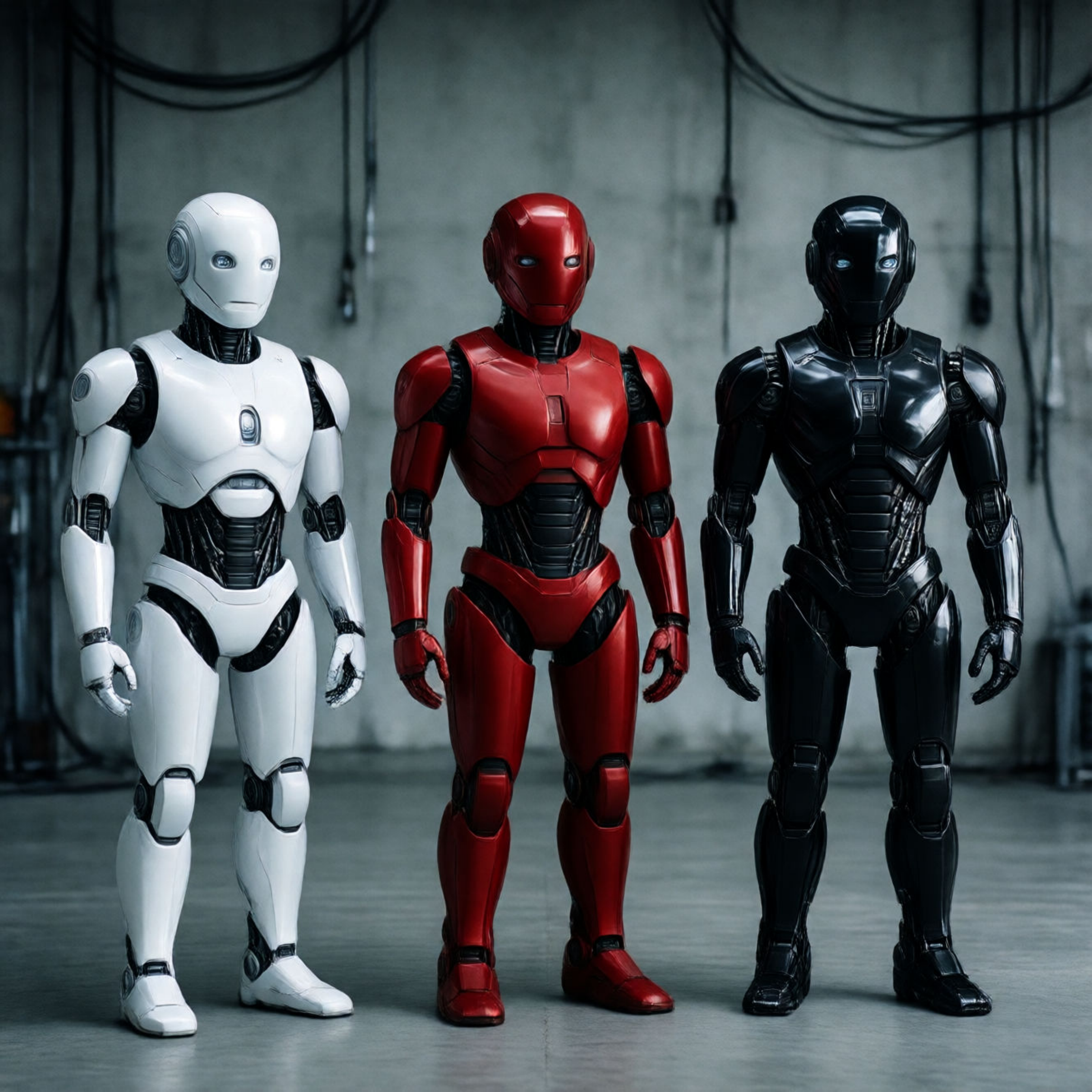} \\[1pt]

        \includegraphics[width=0.192\textwidth]{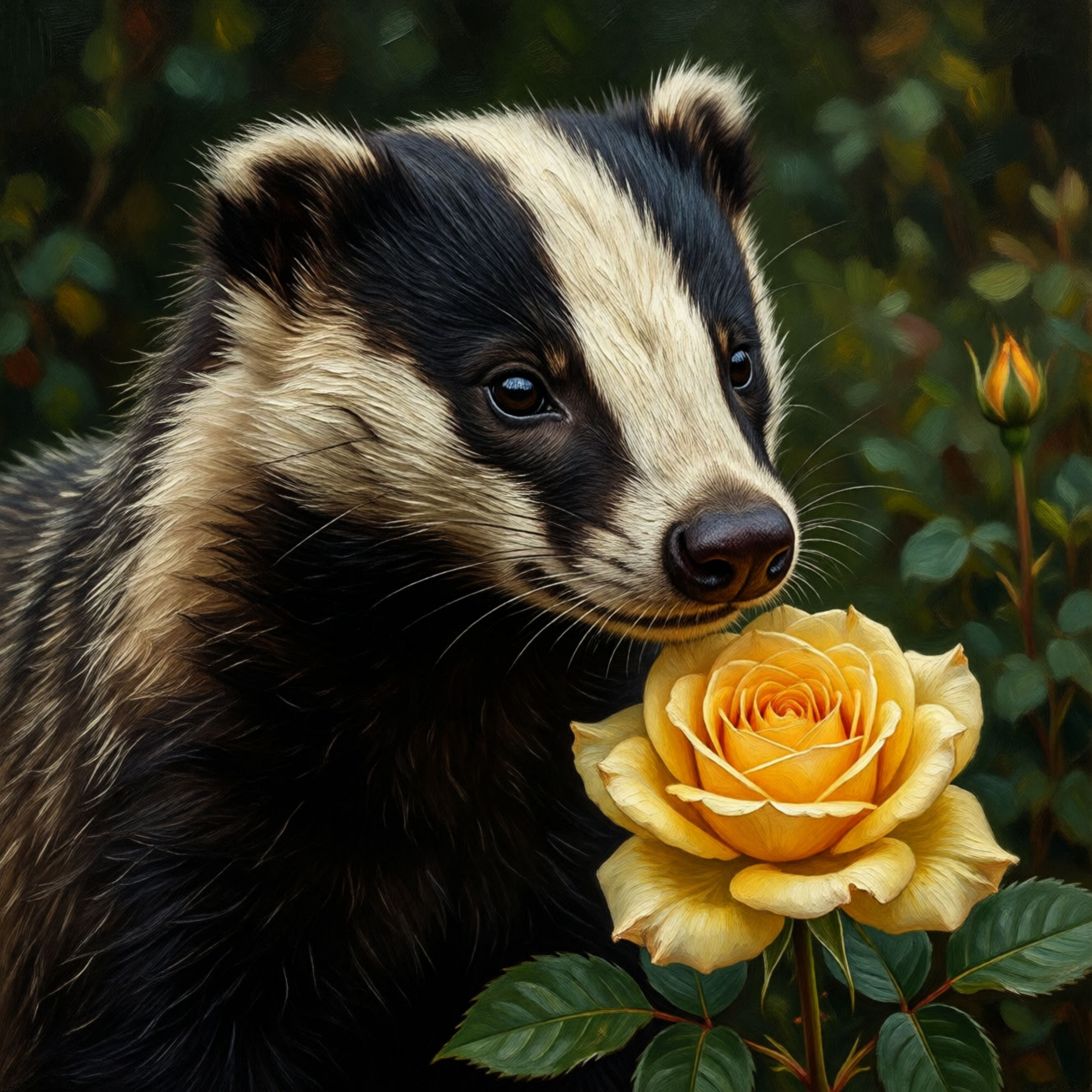} &
        \includegraphics[width=0.192\textwidth]{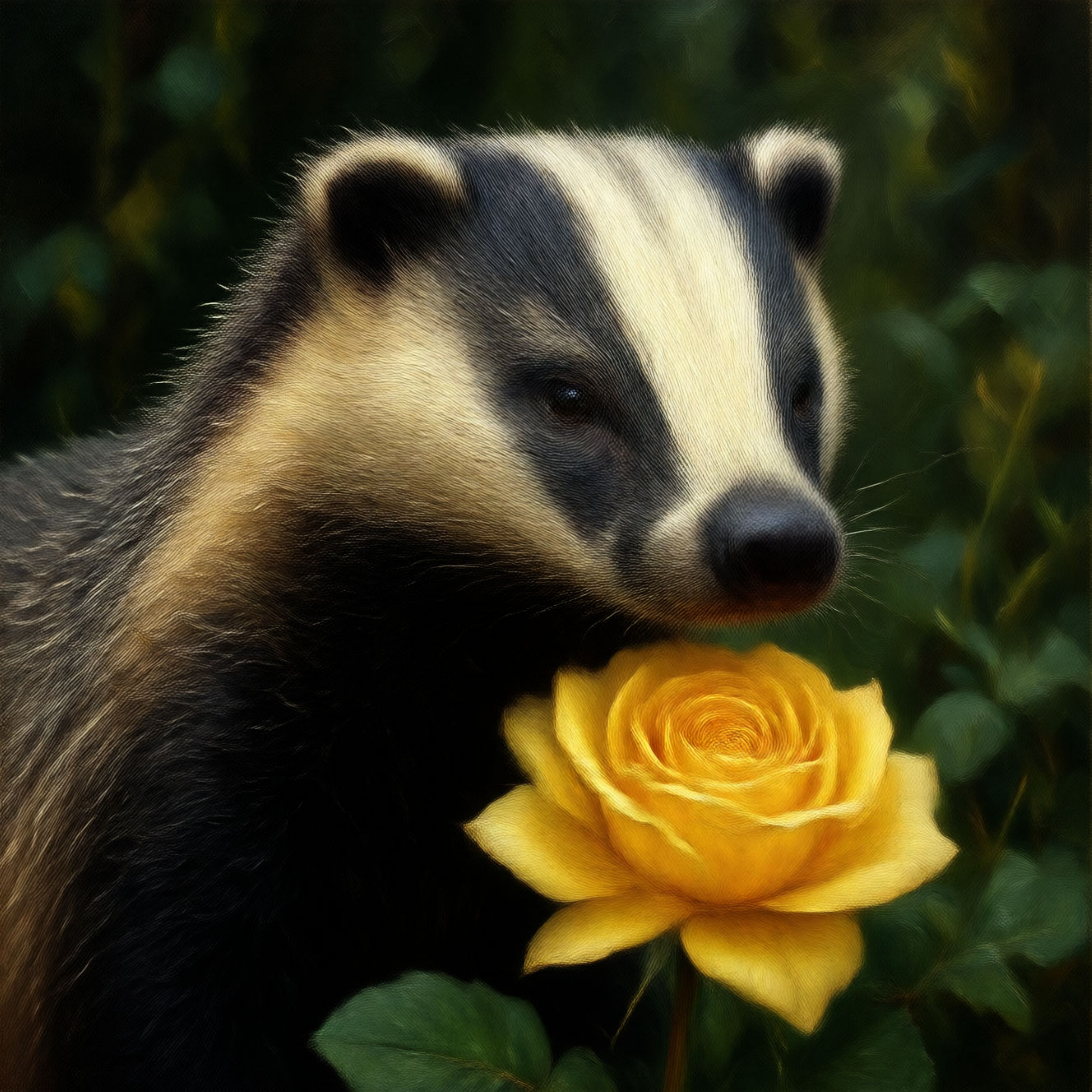} &
        \includegraphics[width=0.192\textwidth]{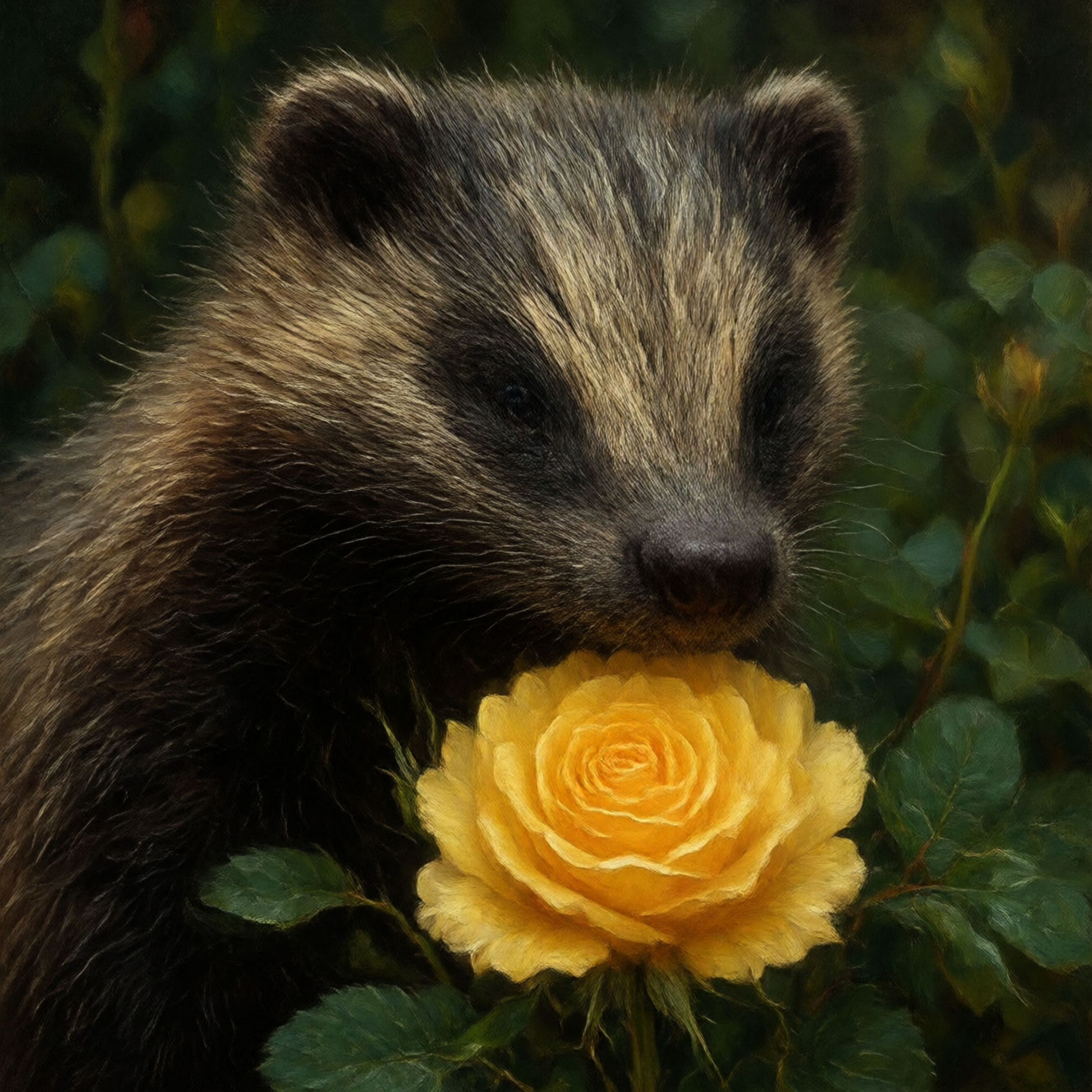} &
        \includegraphics[width=0.192\textwidth]{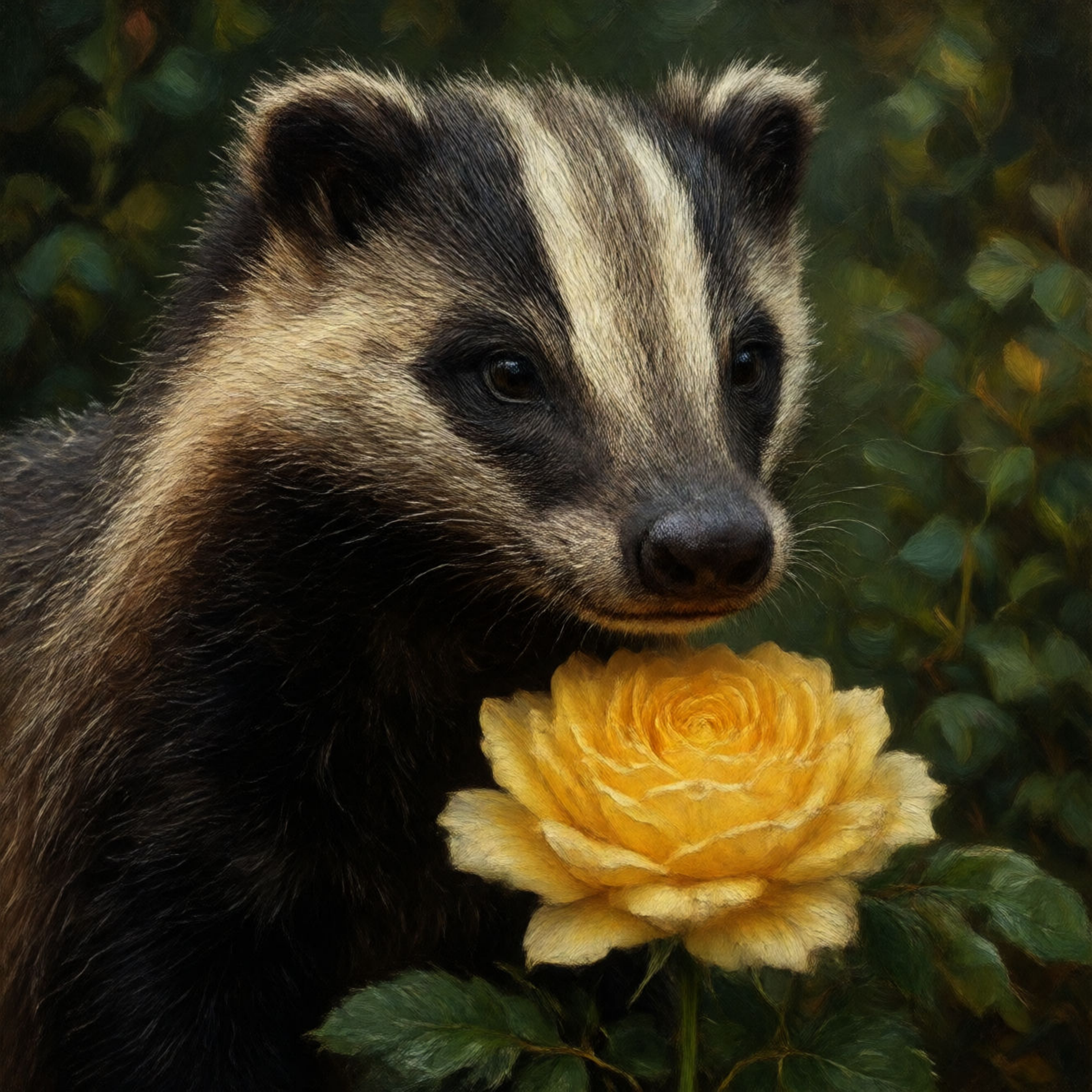} &
        \includegraphics[width=0.192\textwidth]{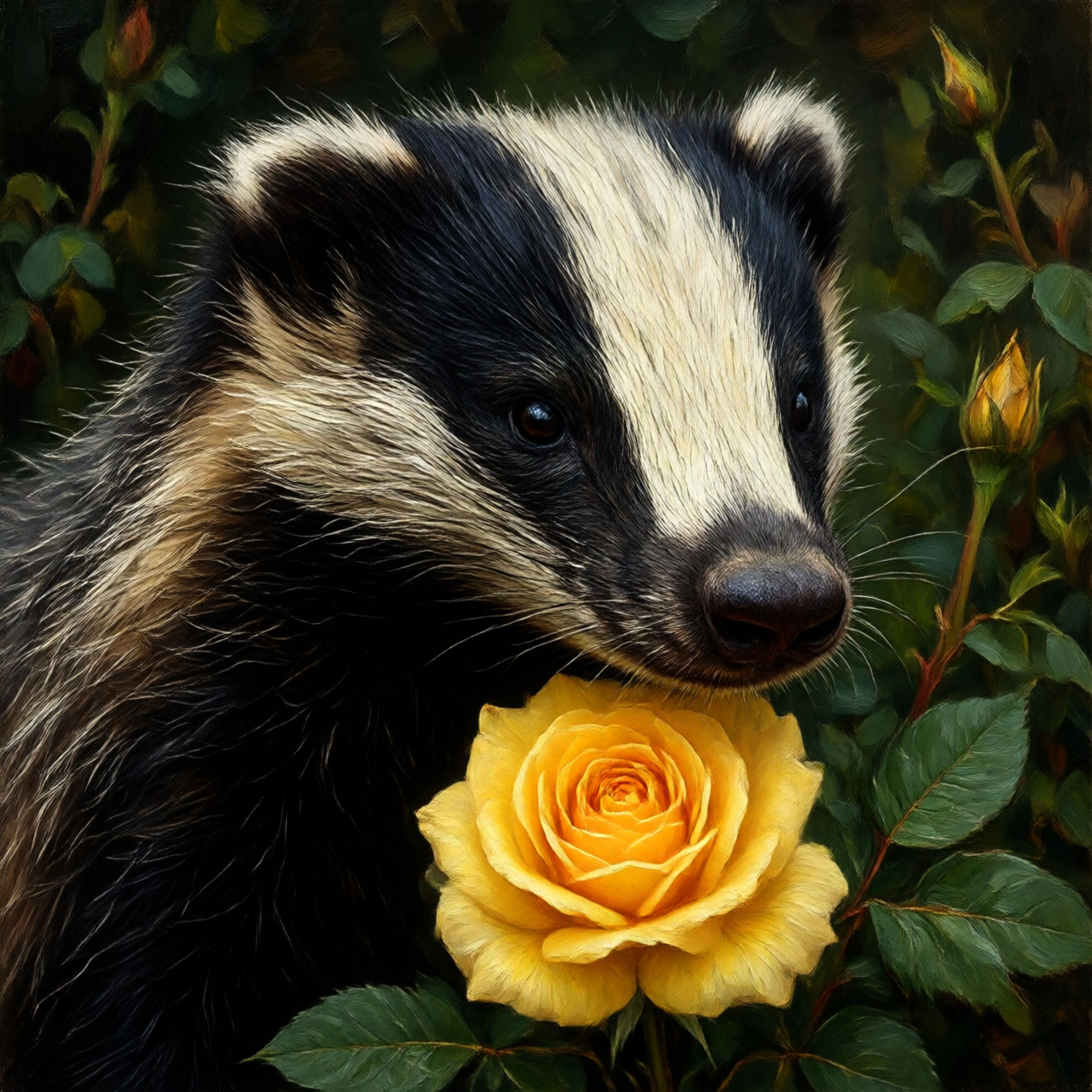}
    \end{tabular}

    \vspace{2pt}
    {\scriptsize
    \textit{Top:} (Composition and Attribute Binding) Three robots with distinct colors and designs.
    \quad
    \textit{Bottom:} (Fine-grained Textures) An oil painting of a badger and a yellow rose.
    }

    \caption{
    Qualitative comparison on Qwen-Image-2512.
    Each row uses the same DPG prompt
    and initial noise across methods. DiT2MoE (Init.) denotes the sparse initialization
    obtained with frozen pretrained FFN weights. The displayed prompt descriptions are abbreviated.
    }
    \label{fig:qualitative_comparison}
\end{figure*}

\section{Experiments}
\label{sec:experiments}

\subsection{Experimental Setup}
\label{sec:experimental_setup}

\paragraph{Models and Setup}
We evaluate DiT2MoE on three pretrained few-step
diffusion models: Qwen-Image-2512~\cite{wu2025qwenimagetechnicalreport}
with a 4-step Lightning sampler~\cite{Qwen-Image-Lightning},
Qwen-Image-2.1-Turbo~\cite{qwen2026qwenimage21}
with 8-step sampling, and
FLUX.1-schnell~\cite{flux2024}
with 4-step sampling.
For Qwen-Image-2512, we convert the image-stream FFNs of MMDiT blocks. For Qwen-Image-2.1-Turbo, we convert FFNs in its single-stream DiT blocks. For FLUX.1-schnell, we convert FFNs in both double-stream and single-stream blocks.
We exclude the first and last blocks from
conversion and leave the remaining architecture
unchanged.
DiT2MoE variants only use text prompts while the Dense2MoE baseline follows its image-based training protocol.

\paragraph{Evaluation} We evaluate generation quality using GenEval~\cite{ghosh2023geneval}, DPG-Bench~\cite{hu2024ella}, CLIP similarity~\cite{radford2021clip}, and ImageReward~\cite{xu2023imagereward}. For each method, we generate 6,472 images using identical evaluation prompts and latent seeds under the same few-step sampling protocol. GenEval and DPG-Bench use their official evaluators on the respective subsets. CLIP and ImageReward reuse these generated images and their prompts, with scores averaged over the combined set.

\paragraph{Baselines and Implementation Details.}
We compare DiT2MoE against the dense teacher,
Dense2MoE~\cite{zheng2025dense2moe}, and a diffusion
adaptation of CMoE~\cite{pei2026analytical}.
For fair comparison, all methods use the same
pretrained backbone, converted FFNs, activation
budget, and evaluation protocol.
We approximately match total training compute
at 400 A800 GPU-hours, including initialization
and subsequent optimization.
DiT2MoE and Dense2MoE require an initial training
stage to construct sparse models, whereas CMoE
uses low-cost analytical initialization.
Dense2MoE retains its original staged distillation
objectives, while CMoE adopts our DMD and LPIPS
recovery losses after expert construction.
DiT2MoE jointly optimizes expert assignments
and routing for 3,000 steps with frozen FFN weights,
followed by 4,000 steps of DMD and LPIPS-based
recovery with five fake-score updates per student
update.
Detailed baseline adaptations and training
configurations are provided in Appendices A and B.

\subsection{Main Results}
\label{sec:main_results}

Table~\ref{tab:main_results} shows that DiT2MoE
outperforms existing MoEfication methods on
11 of 12 metrics across three diffusion models.
On Qwen-Image-2512, it retains 94.9\%--99.8\%
of the dense teacher's performance with only
37.5\% active FFN weights.
Notably, our Stage 1 initialization already preserves substantial
generation quality with frozen FFN weights,
while Stage 2 further improves performance
across all three backbones.
Figure~\ref{fig:qualitative_comparison} illustrates
how Stage 1 preserves overall composition and
Stage 2 restores fine details and visual realism.
The visualized DPG-Bench prompts do not overlap
with the JourneyDB training prompts.

\subsection{Ablation Studies}
\label{sec:ablation}

We conduct controlled ablations on Qwen-Image-2512
at $37.5\%$ active FFN weights.
Unless otherwise specified, all variants share
the same pretrained model, activation budget,
training schedule, and evaluation protocol.

\subsubsection{Stage 1 Ablations}

\paragraph{Expert Assignment}
Table~\ref{tab:stage1_ablation}(a) present an ablation study of expert assignment strategies, while keeping the
shared neurons and routing design unchanged.
Fixed Taylor Partition groups neurons by importance
ranking and trains only the router and timestep
modulation.
DOT-MoE \cite{bamba2026dotmoe} (Disjoint Sinkhorn) learns exclusive expert
assignments through balanced optimal transport,
while Independent Sinkhorn uses the same relaxation
with independent Top-$m$ selection.
Disjoint LML retains exclusive assignments but
replaces Sinkhorn with LML-based optimization.
Our Independent LML combines non-exclusive Top-$m$
selection with independent per-expert relaxation.
Results show that fixed partitioning performs substantially worse
than learned assignments, highlighting the importance of learning expert assignments.
Notably, Independent Sinkhorn underperforms
Disjoint Sinkhorn because its coupled row and
column constraints are incompatible with
non-exclusive forward selection.
In contrast, LML enforces per-expert capacity
without cross-expert competition, aligning the
relaxation with independent selection
and achieving the best performance.

\textbf{Timestep Modulation and Scaling}
Table~\ref{tab:stage1_ablation}(b) evaluates
scaling compensation and timestep modulation
under the same independent LML assignment.
Scaling compensates for the attenuation introduced
by Top-$K$ normalization, while timestep modulation
adapts expert contributions across denoising steps.
Both components individually improve generation
quality, demonstrating that both magnitude compensation and timestep-dependent adaptation are important for faithful sparse conversion.

\begin{table}[t]
\centering
\caption{
Stage~1 ablations on Qwen-Image-2512 at
37.5\% active FFN.
(a) compares expert assignment strategies
under a common routing configuration.
(b) evaluates timestep modulation and expert
scaling with independent LML assignment.
}
\label{tab:stage1_ablation}
\scriptsize
\setlength{\tabcolsep}{3pt}
\renewcommand{\arraystretch}{1.12}
\resizebox{\columnwidth}{!}{%
\begin{tabular}{@{}lcccc@{}}
\toprule
Variant
& GenEval $\uparrow$
& DPG $\uparrow$
& CLIP $\uparrow$
& IR $\uparrow$ \\
\midrule
\multicolumn{5}{@{}l}{\textit{(a) Expert Assignment}} \\
Fixed Taylor Partition
& 0.1393 & 54.13 & 0.2280 & -0.8465 \\
Disjoint Sinkhorn \cite{bamba2026dotmoe}
& 0.5654 & 81.30 & 0.2850 & 0.7406 \\
Independent Sinkhorn
& 0.5571 & 81.25 & 0.2832 & 0.6844 \\
Disjoint LML
& 0.5960 & 82.05 & 0.2849 & 0.7799 \\
\midrule
\textbf{Independent LML (Ours)}
& \textbf{0.6453}
& \textbf{83.89}
& \textbf{0.2890}
& \textbf{0.8813} \\
\midrule
\multicolumn{5}{@{}l}{\textit{(b) Timestep Modulation and Scaling}} \\
Vanilla Top-$K$
& 0.5292 & 80.66 & 0.2811 & 0.6284 \\
Timestep Modulation
& 0.6088 & 83.44 & 0.2864 & 0.8126 \\
Fixed Scaling ($\alpha=K$)
& 0.6101 & 83.54 & 0.2879 & 0.8345 \\
\midrule
\textbf{ Scaling + Modulation (Ours)}
& \textbf{0.6453}
& \textbf{83.89}
& \textbf{0.2890}
& \textbf{0.8813} \\
\bottomrule
\end{tabular}%
}
\end{table}

\subsubsection{Stage~2 Quality Recovery Objectives}
We investigate the contributions of different
distillation objectives under approximately matched
training compute. Variants without DMD use
$5\times$ more training steps to compensate for
the cost of fake-score updates.
As shown in Table~\ref{tab:stage2_ablation},
replacing DMD with velocity MSE substantially
degrades generation quality, highlighting the
limitations of pointwise reconstruction.
Removing LPIPS also leads to consistent degradation,
demonstrating the benefit of complementary
perceptual supervision.
Conversely, adding MSE to DMD and LPIPS provides
no consistent gains and slightly reduces overall
performance.

\begin{table}[t]
\centering
\caption{
Stage~2 loss ablation on Qwen-Image-2512
at 37.5\% active FFN.
All Stage 2 variants use the same Stage~1 initialization,
fixed expert assignments, and training budget.
}
\label{tab:stage2_ablation}
\scriptsize
\setlength{\tabcolsep}{3pt}
\renewcommand{\arraystretch}{1.12}
\resizebox{\columnwidth}{!}{%
\begin{tabular}{@{}lcccc@{}}
\toprule
Training Objective
& GenEval $\uparrow$
& DPG $\uparrow$
& CLIP $\uparrow$
& IR $\uparrow$ \\
\midrule
Stage~1
& 0.6453 & 83.89 & 0.2890 & 0.8813 \\
\midrule
LPIPS + MSE
& 0.6733 & 84.53 & 0.2895 & 0.9877 \\
DMD + MSE 
& 0.7021 & 85.09 & 0.2898 & 1.0196  \\

DMD + MSE + LPIPS
& 0.7250 & 85.38 & \textbf{0.2927} & 1.0690 \\
\midrule
\textbf{DMD + LPIPS (Ours)}
& \textbf{0.7399} & \textbf{85.52}
& 0.2923 & \textbf{1.0701} \\
\bottomrule
\end{tabular}%
}
\end{table}

\subsection{Effect of Stage 1 Initialization}
\label{sec:initialization}

\begin{figure}[t]
    \centering
    \includegraphics[width=0.9\columnwidth]
        {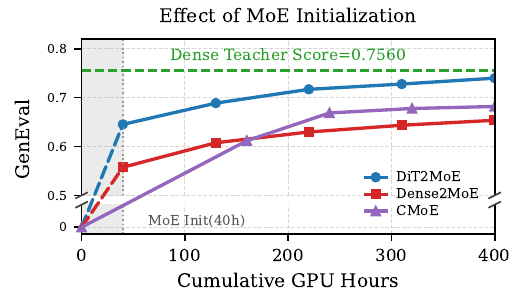}
    \caption{
    Effect of sparse initialization and quality
    recovery on GenEval.
    Cumulative GPU-hours include approximately
    40 hours of initialization for DiT2MoE and
    Dense2MoE, while CMoE uses negligible
    initialization compute.
    }
    \label{fig:initialization_dynamics}
\end{figure}

We investigate how sparse initialization affects
subsequent generation-quality recovery.
For this comparison, DiT2MoE and Dense2MoE
each use approximately 40 GPU-hours for
initialization, followed by their respective
distillation objectives.
In contrast, CMoE requires negligible
initialization compute and uses the same
DMD and LPIPS recovery losses as DiT2MoE.
All methods share the same FFN activation budget,
and we measure cumulative GPU-hours including
both initialization and recovery.
As shown in Figure~\ref{fig:initialization_dynamics},
DiT2MoE achieves higher GenEval than Dense2MoE
immediately after initialization, demonstrating
the effectiveness of our learned sparse structure.
This advantage carries into Stage 2, where
DiT2MoE reaches higher generation quality with
fewer GPU-hours than Dense2MoE.
Furthermore, CMoE remains substantially below
DiT2MoE despite using identical recovery losses,
highlighting the importance of a faithful sparse
initialization.

\subsection{Learned Expert Assignment Analysis}
\label{sec:assignment_analysis}

A natural question is whether allowing neurons
to appear in multiple experts promotes the reuse
of important neurons without excessive redundant
computation.
We measure neuron contributions on dense-teacher
trajectories using the average activation magnitude
weighted by the down-projection norm.
As shown in Table~\ref{tab:assignment_analysis},
neurons assigned to multiple experts exhibit
higher contributions than singly assigned or
unassigned neurons, demonstrating that our method
effectively identifies and preferentially reuses
important neurons.
Importantly, assignment overlap does not
necessarily lead to redundant computation due
to sparse expert routing.
Under learned Top-2 routing, only 9.90\% of
routed slots are duplicated, accounting for
3.30\% of all active FFN slots.
Moreover, duplicated neuron weights are
independently fine-tuned in Stage~2, enabling
expert-specific specialization.
Together with the gains over disjoint assignment
in Table~\ref{tab:stage1_ablation}, these results
support importance-aware neuron reuse with
limited co-activation redundancy.
Further details are provided in Section D of the Appendix.

\begin{table}[t]
\centering
\caption{
Analysis of non-shared neuron assignments.
Statistics are averaged across converted FFN layers.
Contributions are measured on teacher
trajectories and normalized within each layer.
}
\label{tab:assignment_analysis}
\small
\setlength{\tabcolsep}{4pt}
\renewcommand{\arraystretch}{1.12}
\resizebox{\columnwidth}{!}{%
\begin{tabular}{@{}lccc@{}}
\toprule
Membership
& Neuron Ratio (\%)
& Norm. Contrib.
& Contrib. Mass (\%) \\
\midrule
Unassigned ($n_j=0$)
& 52.98 & 0.783 & 41.94 \\
Assigned Once ($n_j=1$)
& 21.43 & 1.097 & 23.54 \\
Overlapping ($n_j\geq 2$)
& 25.59 & \textbf{1.338} & 34.52 \\
\bottomrule
\end{tabular}%
}
\end{table}

\subsection{Shared and Routed Expert Allocation}
In this section, we investigate how to allocate shared and routed
experts under a fixed activation budget and whether conditional routing is necessary.
We compare different $(S,K)$ configurations and
include two routerless baselines:
static Taylor pruning and learned shared-only selection.
The former directly retains the top-ranked neurons
by Taylor importance without retraining, while
the latter learns neuron selection through two-level
LML optimization with channel-wise timestep modulation.
All learned variants use the same Stage~1 training budget.
Among configurations with routed experts, allocating a larger fraction of the fixed activation budget to shared experts improves performance, with \((S,K)=(4,2)\) achieving the best results.
Although learned shared-only selection substantially
outperforms static pruning, our shared-and-routed
design achieves better results, demonstrating
the benefit of combining shared experts
with conditional routing. Further details on shared-only baseline are provided in Appendix C.

\begin{table}[t]
\centering
\caption{
Shared and routed expert allocation on Qwen-Image-2512
at 37.5\% active FFN.
Static Taylor pruning requires no training,
while all learned variants undergo 3,000 steps
of optimization.
}
\label{tab:shared_ablation}
\scriptsize
\setlength{\tabcolsep}{3pt}
\renewcommand{\arraystretch}{1.12}
\resizebox{\columnwidth}{!}{%
\begin{tabular}{@{}lcccc@{}}
\toprule
Configuration
& GenEval $\uparrow$
& DPG $\uparrow$
& CLIP $\uparrow$
& IR $\uparrow$ \\
\midrule
Static Taylor ($S=6$)
& 0.0000 & 9.24 & 0.1669 & -2.2533 \\
Learned Shared-Only ($S=6$)
& 0.6011 & 81.73 & 0.2803 & 0.7913 \\
$S=0,\ K=6$
& 0.5859 & 80.90 & 0.2813 & 0.7538 \\
$S=2,\ K=4$
& 0.6208 & 83.27 & 0.2855 & 0.8697 \\

\midrule
\textbf{$S=4,\ K=2$ (Ours)}
& \textbf{0.6453}
& \textbf{83.89}
& \textbf{0.2890}
& \textbf{0.8813} \\
\bottomrule
\end{tabular}%
}
\end{table}

\subsection{Inference Latency}
We replace the Python dispatch loop with a ScatterMoE grouped-GEMM
backend. The existing router, top-$K$ assignments and expert weights are preserved.
On an NVIDIA A800 GPU in BF16, DiT2MoE with $37.5\%$
active FFN weights achieves \textbf{$1.98\times$} FFN and
\textbf{$1.25\times$} end-to-end speedups at batch size 8
(Table~\ref{tab:ffn_dit_latency}); the corresponding batch-1
speedups are $1.68\times$ and $1.21\times$.
End-to-end FLOPs are reduced by approximately one-third,
while peak memory usage remains comparable to the dense model.
The optimized backend produces numerically equivalent outputs. More details are provided in Section B of the Appendix.

\begin{table}[t]
\centering
\caption{
FFN and end-to-end latency for the $37.5\%$ active-FFN model
($S=4$, $K=2$) on an NVIDIA A800 80\,GB GPU in BF16.
End-to-end latency includes text encoding, the complete four-step denoising trajectory at a resolution of 1328 × 1328, and VAE decoding.
}
\label{tab:ffn_dit_latency}
\scriptsize
\setlength{\tabcolsep}{3pt}
\renewcommand{\arraystretch}{1.12}
\resizebox{\columnwidth}{!}{%
\begin{tabular}{@{}c rrr rrr@{}}
\toprule
\multirow{2}{*}{Batch}
& \multicolumn{3}{c}{FFN Latency (ms)}
& \multicolumn{3}{c}{End-to-End Latency (s)} \\
\cmidrule(lr){2-4}
\cmidrule(l){5-7}
& Dense & Sparse & Speedup $\uparrow$
& Dense & Sparse & Speedup $\uparrow$ \\
\midrule
1 & 4.334  & 2.578  & $1.68\times$
  & 3.218  & 2.652  & $1.21\times$ \\
8 & 34.786 & 17.600 & $\mathbf{1.98\times}$
  & 24.473 & 19.641 & $\mathbf{1.25\times}$ \\
\bottomrule
\end{tabular}%
}
\end{table}

\section{Conclusion}
\label{sec:conclusion}

We presented DiT2MoE, an efficient two-stage framework for converting pretrained dense diffusion transformers into sparse MoE models using only text prompts. Our approach first learns a sparse structure that approximates dense computation through low-cost joint expert assignment and routing optimization, providing a strong initialization for subsequent generative distillation. It then recovers generation quality through student-trajectory supervision and distribution matching. By separating structural initialization from generative quality recovery, DiT2MoE enables efficient dense-to-MoE conversion while largely preserving the capabilities of pretrained diffusion models.

\section{LLM Usage Statement}
\label{sec:llm_usage_statement}
The authors used ChatGPT to assist with code implementation, debugging, language editing, and improvements to manuscript clarity. All research ideas, methodological designs, experimental decisions, and scientific conclusions were developed by the authors. The authors reviewed and revised the LLM-assisted content and take full responsibility for the correctness and integrity of the work.

{
    \small
    \bibliographystyle{ieeenat_fullname}
    \bibliography{main}
}

\newpage
\appendix

\section{Two-Stage Training Procedure}
\label{app:training_procedure}

Algorithm~\ref{alg:dit2moe} summarizes the complete
DiT2MoE training procedure.
Both stages use JourneyDB captions as text prompts
without accessing paired images; all supervision
is derived from model-generated trajectories.
Stage~1 learns a sparse initialization with frozen
pretrained FFN weights, while Stage~2 fixes the
expert structure and recovers generation quality
through student-trajectory distillation.

\begin{algorithm}[t]
\caption{Two-Stage Training of DiT2MoE}
\label{alg:dit2moe}
\scriptsize
\begin{algorithmic}[1]
\Require Dense teacher $v_T$, prompt set $\mathcal C$,
expert configuration $(E,S,K)$
\Ensure Sparse student $v_\theta$

\Statex \textbf{Stage 1: Sparse Initialization}

\State Select shared neurons using Taylor importance
\State Initialize affinities $A$, router, and timestep adapter
\State Freeze pretrained FFN weights

\For{$t=0,\ldots,T_1-1$}
    \State Sample $c\sim\mathcal C$ and
    $z_1\sim\mathcal N(0,I)$
    \State Generate teacher rollout
    $\{(z_{\sigma_i},\sigma_i,c)\}_i$
    \For{each converted FFN}
        \State $P\gets\mathrm{LML}_{\tau_t}(A)$
        \State $M_{je}\gets
        \mathbb I[j\in\mathrm{Top}_m(A_{:,e})]$
        \State $\widetilde M\gets
        P+\mathrm{sg}(M-P)$
        \State Apply scaled Top-$K$ routing and
        timestep modulation
        \State Compute sparse FFN using $\widetilde M$
    \EndFor
    \State Compute $\mathcal L_{\mathrm{MSE}}$
    on teacher rollout states
    \State $\mathcal L_1\gets
    \mathcal L_{\mathrm{MSE}}
    +\lambda_{\mathrm{bal}}\mathcal L_{\mathrm{bal}}
    +\lambda_z\mathcal L_z$
    \State Update $A$, router, and timestep adapter
\EndFor

\Statex \textbf{Stage 2: Generation-Quality Recovery}

\State Fix hard assignments and shared-neuron identities
\State Materialize experts with independent weight copies
\State Unfreeze expert FFN weights;
keep router and timestep adapter trainable
\State Initialize fake-score model $v_\psi\gets v_T$

\For{$t=1,\ldots,T_2$}
    \State Sample $c\sim\mathcal C$,
    $z_1\sim\mathcal N(0,I)$, and step $k$
    \State Roll out student to step $k$;
    detach preceding steps
    \State Compute differentiable clean-latent
    prediction $\hat z_0^\theta$ at step $k$
    \State Sample re-noising level $\rho$
    and $\epsilon\sim\mathcal N(0,I)$
    \State $z_\rho^\theta\gets
    (1-\rho)\hat z_0^\theta+\rho\epsilon$

    \For{$u=1,\ldots,N_{\mathrm{fake}}$}
        \State Update $v_\psi$ via flow matching
        on re-noised student samples
    \EndFor

    \State Compute DMD gradient using frozen
    teacher and fake-score predictions
    \State Compute $\mathcal L_{\mathrm{LPIPS}}$
    at student-visited state $z_{\sigma_k}^\theta$
    \State Update student using DMD, LPIPS,
    and routing regularization
\EndFor

\State \Return $v_\theta$
\end{algorithmic}
\end{algorithm}

\paragraph{Stage~1 optimization.}
We normalize router inputs and initialize the router
and affinity parameters with standard deviations
of $10^{-3}$ and $0.02$, respectively.
The timestep adapter has rank 16.
With pretrained FFN weights frozen, we optimize
the affinities, router, and timestep adapter for
$T_1=3{,}000$ steps.
The learning rates are $3\times10^{-3}$ for the
affinities and adapter and $3\times10^{-4}$ for
the router, with 500 warmup steps followed by
cosine decay.
Load-balancing and router z-loss weights are
$0.1$ and $10^{-3}$, respectively.
Training uses eight A800 GPUs with one sample
per GPU.

\paragraph{Expert materialization.}
During Stage~1, overlapping neuron contributions
are aggregated through the neuron-level gating
formulation in the main paper.
After training, we fix the hard expert assignments
and materialize separate parameter copies for
neurons assigned to multiple routed experts.
These copies share the same pretrained initialization
but are independently fine-tuned in Stage~2.
Consequently, Stage~2 computes sparse FFN outputs
by summing selected expert outputs rather than
aggregating neuron-level gates.

\paragraph{Stage~2 optimization.}
Starting from the initialized sparse model, we
fine-tune expert FFN weights, router, and timestep
adapter for $T_2=4{,}000$ student updates while
keeping expert assignments and shared-neuron
identities fixed.
The learning rates are $10^{-4}$ for FFN weights
and $10^{-5}$ for the router and adapter, with
400 warmup steps.
We use $\lambda_{\mathrm{DM}}=1.0$ and
$\lambda_{\mathrm{LPIPS}}=2.0$, with both routing
regularizers weighted by $10^{-3}$.
The frozen DMD teacher uses CFG scale $3.0$.
The fake-score model performs
$N_{\mathrm{fake}}=5$ updates per student update,
using fresh student samples and a learning rate
of $10^{-4}$.
The re-noising level $\rho$ is sampled from the
student's inference noise schedule independently
of the rollout step.
Both stages use FP32 master weights, FSDP,
and gradient checkpointing.

\section{Implementation Details}
\label{sec:appendix_implementation_details}

\paragraph{Backbone configurations.}
Qwen-Image-2512 has 60 dual-stream blocks, Qwen-Image-2.1-Turbo has
32 single-stream blocks, and FLUX.1-schnell has 19 dual-stream blocks
followed by 38 single-stream blocks. We keep the first and last blocks
of each complete backbone dense, converting the remaining image-stream
FFNs in dual-stream blocks and FFNs in single-stream blocks
(Table~\ref{tab:backbone_config}). All use $(E,S,K)=(16,4,2)$,
activating $37.5\%$ of FFN slots per converted layer.

\begin{table}[t]
\centering
\caption{Numbers of converted FFNs. The first and last blocks of each
complete backbone remain dense.}
\label{tab:backbone_config}
\small
\setlength{\tabcolsep}{3pt}
\begin{tabular}{@{}lccc@{}}
\toprule
Backbone
& \shortstack{Dual-stream\\image FFNs}
& \shortstack{Single-stream\\FFNs}
& Total \\
\midrule
Qwen-Image-2512      & 58 & 0  & 58 \\
Qwen-Image-2.1-Turbo & 0  & 30 & 30 \\
FLUX.1-schnell       & 18 & 37 & 55 \\
\bottomrule
\end{tabular}
\end{table}

\paragraph{Evaluation protocol.}
We generate four images each for 553 GenEval and 1,065 DPG-Bench
prompts, yielding 2,212 and 4,260 images, respectively (6,472 total).
Methods on the same backbone use identical prompts, latent seeds,
sampling schedules, guidance settings, and resolutions.
GenEval and DPG-Bench use their official evaluators; CLIP and
ImageReward are averaged over the combined image set.

\paragraph{Taylor initialization.}
For Qwen-Image-2512, we sample 32 JourneyDB training captions and
run a dense four-step Lightning rollout for each. From each generated
clean latent $z_0$, we draw $\epsilon\sim\mathcal N(0,I)$ and sample
$\sigma$ uniformly from the nonterminal noise levels. The calibration
loss $\mathcal L_b$ is velocity MSE at
$z_\sigma=(1-\sigma)z_0+\sigma\epsilon$, with target $\epsilon-z_0$.

For channel $j$ in layer $\ell$, $\mathcal G_{\ell j}$ contains its
up-projection row and bias and down-projection column. We sum signed
weight--gradient products within each channel before taking the
absolute value and average over $B_{\mathrm{cal}}=32$ batches:
\begin{equation}
T_{\ell j}
= \frac{1}{B_{\mathrm{cal}}}\sum_{b=1}^{B_{\mathrm{cal}}}
  \left|\sum_{i\in\mathcal G_{\ell j}}
    w_i\frac{\partial\mathcal L_b}{\partial w_i}\right|.
\label{eq:shared_taylor_score}
\end{equation}
The $Sm$ highest-scoring channels in each layer become the fixed
shared-neuron set.

\paragraph{LML solver.}
We anneal $\tau$ over 3,000 steps as
$\tau_t=0.05+0.475[1+\cos(\pi t/2999)]$, $t=0,\ldots,2999$.
For each centered affinity column, we solve $\sum_j P_{je}=m$
using 64 float64 bisection steps, with bounds given by the column
extrema shifted by $-\tau\log(m/(N-m))$.
Writing $p_j=P_{je}$, $g_j=\partial\mathcal L/\partial p_j$,
and $w_j=p_j(1-p_j)$, implicit differentiation yields
\begin{equation}
\frac{\partial\mathcal L}{\partial A_{je}}
=\frac{w_j}{\tau}
\left(g_j-\frac{\sum_i w_i g_i}{\sum_i w_i}\right).
\end{equation}

\paragraph{Disjoint LML ablation.}
Unlike independent Top-$m$ selection, the disjoint variant uses
deferred acceptance on affinity scores: neurons propose to their
best available expert, which retains the highest-scoring proposals
up to capacity. Each non-shared neuron is assigned once and each
expert receives $m$ neurons. Both variants use the same soft LML
projection for straight-through gradients.

\paragraph{Dense2MoE baseline.}
We fuse the four-step Lightning weights into Qwen-Image-2512 before
applying Dense2MoE~\cite{zheng2025dense2moe}. To match conversion
scope, we omit MoB, AdaLN compression, and dense-tail inference.
After excluding benchmark prompt matches, we split 20,000 JourneyDB
image--caption pairs into 19,487 training, 256 calibration, and 257
validation pairs. We compute flow-matching gradients on calibration
images and sum absolute weight--gradient products per channel.
Outside shared experts, channels are sorted by score and divided
into contiguous equal-sized groups with fixed assignments.
We allocate approximately 40 A800 GPU-hours to shared-expert
enhancement and, after accounting for calibration, use the remaining
budget for full MoE distillation with output and dynamically weighted
feature matching.

\paragraph{CMoE baseline.}
For CMoE~\cite{pei2026analytical}, we collect post-GELU rather than
SwiGLU activations across noise levels. Shared neurons are selected
by activation frequency; remaining neurons are clustered by
co-activation with equal capacities, and representative neuron
activations provide routing scores. We then fine-tune expert weights
with the same DMD and LPIPS losses as DiT2MoE, keeping assignments
and analytical routing fixed.

\paragraph{Initialization comparison.}
For Fig.~4 of the main paper, cumulative GPU-hours include
initialization and recovery. DiT2MoE's Stage~1 and Dense2MoE's
shared-expert enhancement each use approximately 40 A800 GPU-hours;
Dense2MoE subsequently follows its full MoE distillation recipe.

\paragraph{Latency measurement.}
The dense FFN uses PyTorch linear layers with GELU. The sparse FFN
uses standard GEMMs for shared experts, ScatterMoE Triton grouped
GEMMs for routed experts, and a fused Triton merge.
Full-FFN timing includes routing, dispatch, expert computation, and
merging with output bias; fixed-routing timing precomputes routing
indices and gates but still includes dispatch and merging.
After 30 warmup forwards, CUDA events time 100 forwards per repeat,
with synchronization at loop boundaries. We report the median of five
per-forward averages. DiT-only latency uses 2 warmup and 5 measured
runs with synchronized CUDA events. Full-pipeline latency uses
1 warmup and 3 measured runs with synchronized wall-clock timing,
including text encoding and VAE decoding. Both report medians.
Table~6 in the main paper uses full-FFN timing for FFN speedups.

\section{Additional Ablations}
\label{app:additional_ablations}

\paragraph{Stage~2 recovery objectives.}
The variants in Table~3 of the main paper share the Stage~1 checkpoint,
fixed assignments, and activation budget. For LPIPS + MSE, which has
no fake-score updates, we use five times as many student updates to
approximately match compute. Included loss terms have weights
$\lambda_{\mathrm{DM}}=1.0$, $\lambda_{\mathrm{LPIPS}}=2.0$, and
$\lambda_{\mathrm{MSE}}=1.0$. MSE-based ablations reconstruct the
unguided teacher's velocity on student-visited states:
\begin{equation}
\mathcal{L}_{\mathrm{MSE}}^{(2)}
=\mathbb{E}_{\mathcal D_\theta}\!\left[
\left\|v_\theta(z_\sigma,\sigma,c)
-\operatorname{sg}\bigl(v_T^{u}(z_\sigma,\sigma,c)\bigr)
\right\|_2^2\right],
\end{equation}
where $v_T^{u}$ is the frozen unguided teacher.
The final DiT2MoE objective excludes MSE.

\begin{table}[!htbp]
\centering
\caption{FFN activation-budget comparison on Qwen-Image-2512
with $E=16$. The $37.5\%$ row reproduces the Stage~2 result in
Table~1 of the main paper.}
\label{tab:activation_budget}
\small
\setlength{\tabcolsep}{3pt}
\begin{tabular}{@{}lccccc@{}}
\toprule
Active FFN & $(S,K)$ & GenEval & DPG & CLIP & IR \\
\midrule
$25\%$ & $(2,2)$ & 0.6833 & 84.31 & 0.2892 & 0.9864 \\
$37.5\%$ (Ours) & $(4,2)$ & 0.7399 & 85.52 & 0.2923 & 1.0701 \\
$50\%$ & $(4,4)$ & 0.7436 & 85.88 & 0.2932 & 1.1052 \\
\bottomrule
\end{tabular}
\end{table}

\paragraph{Shared-neuron selection.}
Replacing Taylor importance with the dense-FFN contribution score
reduces all four metrics (Table~\ref{tab:shared_init}). Both variants
use the same Stage~1 setup with fixed shared-neuron identities.

\begin{table}[!htbp]
\centering
\caption{Stage~1 shared-neuron selection on Qwen-Image-2512.
$\Delta$ denotes contribution-score minus Taylor-importance results.}
\label{tab:shared_init}
\small
\setlength{\tabcolsep}{3pt}
\begin{tabular}{@{}lcccc@{}}
\toprule
 & GenEval & DPG & CLIP & IR \\
\midrule
Taylor importance & 0.6453 & 83.89 & 0.2890 & 0.8813 \\
Contribution score & 0.6175 & 82.21 & 0.2854 & 0.7997 \\
$\Delta$ & $-0.0278$ & $-1.68$ & $-0.0036$ & $-0.0816$ \\
\bottomrule
\end{tabular}
\end{table}

\paragraph{Shared-only and disjoint expert baselines.}
Table~5 of the main paper compares static Taylor pruning,
learned shared-only selection, and routed allocations at $37.5\%$
activation. The learned shared-only variant uses two-level LML
optimization with channel-wise timestep modulation; routed variants
use independent expert assignments.

\section{Expert Assignment Statistics}
\label{sec:supp_assignment}

\paragraph{Contribution by assignment group.}
For Table~4 of the main paper, we measure non-shared neuron
contributions over four dense-teacher denoising steps on 32 GenEval
prompts, normalized by the mean contribution of non-shared neurons
within each layer. We group neurons by assignment count
$n_{\ell j}=\sum_{e=1}^{R}M_{\ell je}$; contribution mass is each
group's share of the non-shared total. Statistics are computed per
layer and then averaged, so products of reported group ratios and
mean contributions need not reproduce averaged contribution masses.

\paragraph{Co-activation overlap.}
On the same trajectories, we apply the Stage~1 router
$(E,S,K)=(16,4,2)$. For selected routed expert neuron sets $A$ and
$B$, the duplicate fraction of routed slots is
$\mathbb E[|A\cap B|/(2m)]$; multiplying by $K/(S+K)$ gives the
fraction among all active FFN slots.

\section{Qualitative Results}
\label{app:qualitative_results}

Figure~3 of the main paper uses the same full DPG-Bench prompt and
initial noise within each row; the displayed descriptions are
abbreviated. Stage~1 preserves the robots' color assignments and the
badger's flower placement, while Stage~2 improves fine details.
The full generation prompts are given below.

\paragraph{Badger example.}
\begin{quote}
\small
An intricately detailed oil painting captures the essence of
a young badger, its fur rendered with fine brushstrokes that
give it a tactile quality. The badger's snout is gently poised
above a vibrant yellow rose, which stands out against a
backdrop of muted green foliage. The contrast between the
animal's coarse fur and the delicate petals of the rose is
emphasized through the artist's skillful use of texture and color.
\end{quote}

\paragraph{Robots example.}
\begin{quote}
\small
Three robots of varying hues are positioned in a row, each
with a distinct and sleek design. The white robot on the left
boasts a glossy finish and rounded edges, while the central
red robot has a more angular form with matte surfaces.
To the right, the black robot stands with a metallic sheen,
its articulated joints suggesting advanced mobility.
They are all placed on a smooth, gray concrete floor,
and behind them is a plain wall with a few cables hanging
neatly arranged.
\end{quote}

\end{document}